\documentclass{article}

\usepackage{microtype}
\usepackage{graphicx}
\usepackage{subfigure}
\usepackage{booktabs} 

\usepackage{hyperref}

\usepackage{rotating}
\usepackage{booktabs}
\usepackage{amsfonts}       
\usepackage{nicefrac}       
\usepackage{xcolor}         
\usepackage{multirow}
\usepackage{bbm}
\usepackage{natbib}
\usepackage{tikz}
\usepackage{amssymb}
\usetikzlibrary{arrows}
\usepackage{adjustbox}
\usepackage{latexsym}
\usepackage{epsfig}
\usepackage{fancybox}
\usepackage{pstricks}
\usepackage{caption}
\usepackage{fancyhdr}
\usepackage{color}
\usepackage{wrapfig,lipsum}
\usepackage{cancel}
\usepackage{kotex}
\usepackage{framed}
\usepackage{amsthm}

\def\qed{\space$\square$ \par \vspace{.15in}}
\def\hat{\widehat}

\newcommand{\bc}{\begin{center}}
\newcommand{\ec}{\end{center}}
\newcommand{\be}{\begin{equation}}
\newcommand{\ee}{\end{equation}}
\newcommand{\ba}{\begin{array}}
\newcommand{\ea}{\end{array}}
\newcommand{\bean}{\begin{eqnarray*}}
\newcommand{\eean}{\end{eqnarray*}}
\newcommand{\bea}{\begin{eqnarray}}
\newcommand{\eea}{\end{eqnarray}}
\newcommand{\ben}{\begin{enumerate}}
\newcommand{\een}{\end{enumerate}}
\newcommand{\bed}{\begin{itemize}}
\newcommand{\eed}{\end{itemize}}
\definecolor{mmd}{HTML}{D2302C}
\definecolor{laftr}{HTML}{A04EF6}
\definecolor{sipm}{HTML}{F95700}
\definecolor{ftm}{HTML}{343148}

\usepackage[accepted]{icml2023}

\usepackage{amsmath}
\usepackage{amssymb}
\usepackage{mathtools}
\usepackage{amsthm}

\usepackage[capitalize,noabbrev]{cleveref}

\definecolor{purple}{RGB}{112,48,160}
\definecolor{green}{RGB}{56,87,35}
\definecolor{red}{RGB}{255,0,0}

\def\qed{\space$\square$ \par \vspace{.15in}}
\def\hat{\widehat}
\newcommand{\E}{\mbox{{\rm E}}}
\theoremstyle{plain}
\newtheorem{theorem}{Theorem}[section]
\newtheorem{proposition}[theorem]{Proposition}
\newtheorem{lemma}[theorem]{Lemma}

\theoremstyle{definition}

\newtheorem{assumption}[theorem]{Assumption}
\theoremstyle{remark}
\newtheorem{remark}[theorem]{Remark}

\usepackage[textsize=tiny]{todonotes}

\icmltitlerunning{ALTBI: Constructing Improved Outlier Detection Models via Optimization of Inlier-Memorization Effect}

\begin{document}

\twocolumn[
\icmltitle{ALTBI: Constructing Improved Outlier Detection Models via Optimization of Inlier-Memorization Effect}



\icmlsetsymbol{equal}{*}

\begin{icmlauthorlist}
\icmlauthor{Seoyoung Cho}{sungshin1}
\icmlauthor{Jaesung Hwang}{skt}
\icmlauthor{Kwan-Young Bak}{sungshin1,sungshin2}
\icmlauthor{Dongha Kim}{sungshin1,sungshin2}

\end{icmlauthorlist}
\icmlaffiliation{sungshin1}{Department of Statistics, Sungshin Women's University, Seoul, Republic of Korea}
\icmlaffiliation{sungshin2}{Data Science Center, Sungshin Women's University, Seoul, Republic of Korea}
\icmlaffiliation{skt}{SK Telecom, Seoul, Republic of Korea}

\icmlcorrespondingauthor{Dongha Kim}{dongha0718@gmail.com}


\vskip 0.3in
]



\printAffiliationsAndNotice{}  

\begin{abstract}
Outlier detection (OD) is the task of identifying unusual observations (or outliers) from a given or upcoming data by learning unique patterns of normal observations (or inliers). 
Recently, a study introduced a powerful unsupervised OD (UOD) solver based on a new observation of deep generative models, called \textit{inlier-memorization (IM) effect}, which suggests that generative models memorize inliers before outliers in early learning stages. 
In this study, we aim to develop a theoretically principled method to address UOD tasks by \textit{maximally utilizing the IM effect}.  
We begin by observing that the IM effect is observed more clearly when the given training data contain fewer outliers. 
This finding indicates a potential for enhancing the IM effect in UOD regimes if we can effectively exclude outliers from mini-batches when designing the loss function.   
To this end, we introduce two main techniques: 1) increasing the mini-batch size as the model training proceeds and 2) using an adaptive threshold to calculate the truncated loss function. 
We theoretically show that these two techniques effectively filter out outliers from the truncated loss function, allowing us to utilize the IM effect to the fullest. 
Coupled with an additional ensemble technique, we propose our method and term it \textit{Adaptive Loss Truncation with Batch Increment (ALTBI)}.   
We provide extensive experimental results to demonstrate that ALTBI achieves state-of-the-art performance in identifying outliers compared to other recent methods, even with lower computation costs.
Additionally, we show that our method yields robust performances when combined with privacy-preserving algorithms. 

\end{abstract}

\section{Introduction}
\label{sec:intro}

\paragraph{Outlier detection:}
Outlier detection (OD) is an important task in various fields, aiming to identify unusual observations, or outliers. 
This process involves learning the distinct patterns of normal observations, known as inliers, and developing efficient scores to distinguish between inliers and outliers. 
The ability to accurately detect outliers is essential for ensuring data quality and reliability in applications such as fraud detection, network security, and fault diagnosis. 

OD tasks can be divided into three cases depending on availability of anomalousness information of given training data. 
Supervised OD (SOD) uses labeled data to classify each sample as either outlier or not. 
Semi-supervised OD (SSOD), also known as out-of-distribution detection, assumes that all training data are normal and builds models based only on these inliers.
Unsupervised OD (UOD) works with data that may contain outliers but has no labels to identify them. 
In UOD tasks, the primary goal is to accurately identify outliers in a given training dataset. 
In general, many real-world anomaly detection tasks belong to UOD because outliers in large datasets are usually unknown beforehand. 
Hence, we focus on addressing UOD problems in this study. 

Recent advancements in machine learning domains have introduced powerful UOD methods. 
In particular, many studies have leveraged deep generative models (DGM) to develop unique scores to identify outliers.
Surprisingly, conventional likelihood was not utilized, as it has been widely recognized that it often confuses outliers with inliers when the models are fully trained \citep{DBLP:conf/iclr/NalisnickMTGL19,https://doi.org/10.48550/arxiv.1906.02994,DBLP:journals/entropy/LanD21}.

Recently, a notable study has highlighted the potential of the likelihood values of DGMs in UOD tasks, based on the observation of the \textit{inlier-memorization (IM) effect} \citep{kim2024odimoutlierdetectionlikelihood}. 
This effect suggests that when a DGM is trained, the loss values, i.e., the negative log-likelihoods, of inliers decrease before those of outliers in early training stages. 
This implies that the likelihood value itself can be a favorable measure to identify outliers with \textit{under-fitted DGMs}.
Leveraging this phenomenon, \citet{kim2024odimoutlierdetectionlikelihood} developed a UOD solver called ODIM, which has demonstrated powerful yet computationally efficient performance in identifying inliers within a dataset.

\begin{figure}[t]
\centering
\includegraphics[width=0.98\columnwidth]{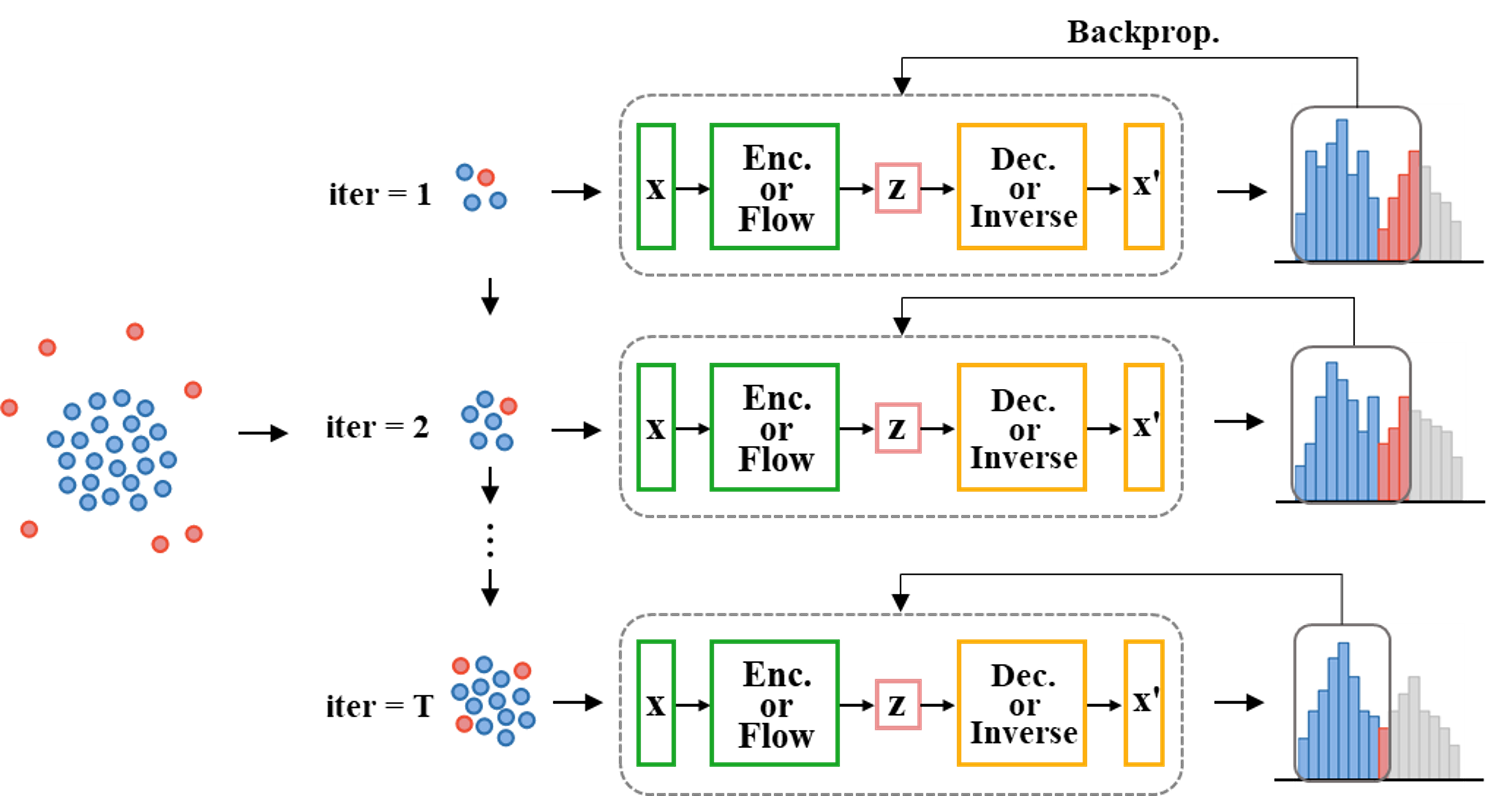} 
\caption{An illustration of ALTBI.}
\label{fig:altbi}
\end{figure}

\paragraph{Improvement of IM effect:}
Inspired by ODIM, this study aims to develop an enhanced UOD method by maximally exploiting the IM effect.  
We begin by observing that the clarity of the IM effect becomes more pronounced as the training data contain fewer outliers. 
We visually illustrate this simple but important finding in Figure \ref{fig:outlier_ratio}. 
This observation suggests that boosting the IM effect could be achieved if we can  effectively separate outliers from inliers and exclude them when constructing loss functions during the early training stages.

To this end, we introduce two key techniques: 1) gradually increasing the mini-batch size and 2) adopting an adaptive threshold to truncate the loss function. 
These techniques are designed to maximize the utility of the IM effect, ensuring more accurate identification of outliers. 
We provide theoretical results showing that these two techniques result in the ratio of outliers included in the truncated loss function decreasing toward zero as training proceeds. 

Additionally, we incorporate an ensemble strategy of loss values within a single DGM with various updates to further enhance our method's performance and stability.
We demonstrate that this simple technique significantly improves the outlier detecting performance without additional computational or resource costs.  

Combining all the above elements, we develop a powerful framework for addressing UOD tasks and term it \textit{Adaptive Loss Truncation with Batch Increment (ALTBI)}. 
We present Figure \ref{fig:altbi} to visualize our method.  
Our method has several notable advantages over other existing UOD solvers.
First, our method consistently achieves superior results in detecting outliers from a given dataset. 
Through extensive experiments analyzing 57 datasets, we demonstrate that ALTBI achieves state-of-the-art performance in outlier detection. 

Additionally, ALTBI only requires a simple and under-fitted likelihood-based DGM, training with variational autoencoders (VAE, \citet{kingma2013auto}) or normalizing flows (NF, \citet{DBLP:conf/iclr/DinhSB17}) for up to hundreds of mini-batch updates.  
This makes ALTBI highly efficient, requiring significantly reduced computational costs compared to other recent methods. 
Our findings indicate that ALTBI is a promising approach for efficient and effective UOD solution in various practical applications.

The reminder of our paper is organized as follows. 
We first provide a brief review of related research on OD problems, primarily focusing on SSOD and UOD. 
Then, we offer detailed descriptions of ALTBI along with its motivations, followed by its theoretical discussions. 
The results of various experiments are presented, including performance tests, ablation studies, and further discussions related to data privacy. 
Finally, concluding remarks are provided.
The key contributions of our work are:
\bed
\item We find that the IM effect is observed more apparently when the training data have fewer outliers. 
\item We develop a theoretically well-grounded and powerful UOD solution called ALTBI, using truncated loss functions with incrementally increasing mini-batch sizes.
\item We empirically validate the superiority of ALTBI in detecting outliers by analyzing 57 datasets. 
\eed

\section{Related works}
\label{sec:review}

We first review studies dealing with SSOD problems. 
SVDD \citep{svdd} uses kernel functions to construct a boundary around a dataset for outlier detection, while DeepSVDD \citep{deepsvdd} extends SVDD by using a deep autoencoder to obtain a feature space where normal data lies within a boundary, while anomalies fall outside. 
DeepSAD \citep{deepsad} generalizes DeepSVDD by considering an extended scenarios where a small amount of labeled outliers are also available.

Self-supervised learning has been widely applied to address SSOD tasks \citep{csi,golan2018deep}. 
In particular, SimCLR \citep{simclr} leverages the contrastive learning to obtain high-quality feature representations of inliers, and ICL \citep{icl} maximizes mutual information between masked and unmasked parts of data to successfully detect outliers.

Numerous traditional approaches have been proposed to address UOD problems. 
LOF \citep{10.1145/335191.335388} detects local outliers in a dataset based on density. 
This idea has been extended to CBLOF \citep{he2003discovering}, which identifies outliers based on the distance from the nearest cluster and the size of the cluster it belongs to, measuring the significance of an outlier. 
MCD \citep{fauconnier2009outliers} identifies outliers by finding a subset of the data with the smallest covariance determinant, providing powerful outlier scores using estimates of location and scatter. 
IF \citep{liu2008isolation} detects anomalies by isolating data points using tree structures. 

There are also various techniques to solve UOD problems using deep learning models. RDA \citep{rda} extends deep autoencoder by incorporating robustness against outliers, and DSEBM \citep{zhai2016deep} generates an energy function as the output of a deterministic deep neural network. Additionally, ODIM \citep{kim2024odimoutlierdetectionlikelihood} utilizes the inlier-memorization effect observed in the early training updates of deep generative models to identify outliers. DTE \citep{DBLP:journals/corr/abs-2305-18593} estimates the distribution over diffusion time for a given input and uses the mode or mean of this distribution as the anomaly score. 
\section{Detailed description of ALTBI}
\label{sec:method}
\subsection{Preliminaries}
\label{sec:pre}
\paragraph{Notations and definitions}
We introduce notations and definitions frequently used throughout this paper. 
Let $X_1,\ldots,X_n(\in\mathcal{X}\subset{\mathbb{R}^D}) {\sim}P_*$ be $n$ independent random input vectors following the true distribution $P_*$. 
Since training data contain outliers as well as inliers in the UOD regime, we assume that $P_*$ is a mixture of two distributions, i.e., $P_* = (1-\alpha)P_{i}+\alpha P_{o}$, where $P_{i}, P_{o}$ represent the inlier and outlier distributions, respectively, and $\alpha\in(0,1)$ is the outlier ratio. 
And we define the support of $P_i$ and $P_o$ as $\mathcal{X}_i$ and $\mathcal{X}_o$, respectively (hence $\mathcal{X}=\mathcal{X}_i\cup\mathcal{X}_o$). 
Since inliers and outliers do not share their supports, we can obviously assume $\mathcal{X}_i\cap \mathcal{X}_o=\emptyset.$

We denote a training dataset comprising $n$ observations by $\mathcal{D}^{\text{tr}}=\{\boldsymbol{x}_1,\ldots,\boldsymbol{x}_n\}$. 
For a given sample $\boldsymbol{x}$, a per-sample loss function with a given DGM is defined as $l(\theta;\boldsymbol{x})$, where $\theta\in\Theta$ represents the parameters for constructing the DGM. 
Since we consider likelihood-based DGMs such as VAE-based ones \citep{kingma2013auto,DBLP:journals/corr/BurdaGS15,kim2020casting}, or NF-based ones \citep{DBLP:journals/corr/DinhKB14,DBLP:conf/iclr/DinhSB17,DBLP:conf/nips/KingmaD18}, $l(\boldsymbol{x};\theta)$ would be the negative log-likelihood (or ELBO). 
Without loss of generality, we assume that $l(\theta;\boldsymbol{x})$ is differentiable and bounded by $[0,1]$ for any $\boldsymbol{x}\in\mathcal{X}$ and $\theta\in\Theta$. 

The risk function calculated over a distribution $P$ is denoted by $L(\theta,P)={E}_{X\sim P}\left[ l(\theta;X) \right]$.  

We respectively abbreviate $L(\theta,P_i)$ and $L(\theta,P_o)$ as $L_i(\theta)$ and $L_o(\theta)$. 
Finally, we denote the minimizer of the inlier risk as $\theta_*$, i.e., $\theta_*=\text{argmin}_{\theta} L_i(\theta)$. 
We assume $L_i(\theta_*)=0$.

\paragraph{Brief review of ODIM}
ODIM \citep{kim2024odimoutlierdetectionlikelihood} is a UOD solver that utilizes the IM effect for the first time. 
The IM effect refers to a phenomenon where, when training a DGM with a given dataset that may contain outliers, the model eventually learns all the patterns of the data, but there is a gap in the memorization order between inliers and outliers. 
That is, the model memorizes inliers first during the early updates. 
Inliers are more prevalent and densely distributed than outliers, thus, reducing the per-sample loss values of inliers first is a more beneficial direction to minimize the overall (averaged) loss function in the early training stages, which might be an intuitive explanation of the IM effect.

ODIM trains likelihood-based DGMs, such as VAE \citep{kingma2013auto}, for a certain number of updates and uses the per-sample loss values as the outlier score. 
To find the optimal number of updates where the IM effect is most clearly observed, ODIM examines the degree of bi-modality in the per-sample loss distribution at each update.

To this end, at each update, ODIM fits a 2-cluster Gaussian mixture model to the per-sample loss distribution and calculates the dissimilarity between the two clusters using measures such as the Wasserstein distance. 
The bi-modality measures are monitored for all updates, and the update with the maximum measure is chosen as the optimal point.

To enhance outlier identification performance, an ensemble technique is also adopted. 
Multiple under-fitted DGMs are independently trained, and the final score of a given sample $\boldsymbol{x}$ is calculated as the average per-sample loss value, given as:
\begin{equation*}
s^{\text{ODIM}}(\boldsymbol{x})=\frac{1}{B}\sum_{b=1}^B l(\theta^{(b)};\boldsymbol{x}),
\end{equation*}
where $\theta^{(b)},b=1,\ldots,B$ are the $B$ estimated parameters from independent initializations.  
And the sample $\boldsymbol{x}$ is regarded as an outlier if its score is large, and vice versa.

\subsection{Relationship between IM effect and outlier ratio}

The IM effect implies that distinguishing inliers from outliers is viable by using the per-sample loss values with an under-fitted DGM. 
In this section, we claim that the clarity of the IM effect increases as the proportion of outliers in the training dataset decreases. 
To demonstrate this, we conduct a simple experiment analyzing the \texttt{PageBlocks} dataset with various outlier ratios ranging from 1\% to 9\%. 
In each setting, we train a VAE for 100 mini-batch updates with a mini-batch size of 128 and evaluate the AUC values of the training data for identifying outliers using per-sample loss values.

\begin{figure}[t]
\centering
\includegraphics[width=0.5\columnwidth]{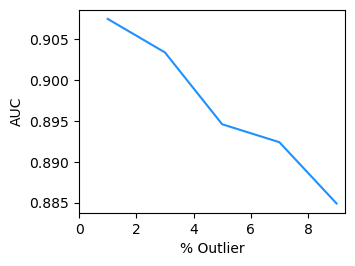} 
\caption{Relationship between the outlier ratio in training data and IM effect.}
\label{fig:outlier_ratio}
\end{figure}

Figure \ref{fig:outlier_ratio} illustrates that the IM effect is observed more clearly as the training data contain fewer outliers, which strongly validates our claim. 
This observation leads to a new belief: if we can effectively filter out outliers when constructing loss values using a mini-batch to update parameters, we can strengthen the IM effect, thereby enhancing outlier detection performance. 
This is the main motivation of our study.

\subsection{Proposed method}
\paragraph{Mini-batch increment and adaptive threshold}
We again note that the goal is to maximize the utility of the IM effect in early training updates. 
To this end, we introduce two simple but powerful strategies to obtain a refined loss function: \textit{1) using a mini-batch size that gradually increases as training proceeds and 2) utilizing a truncated loss function with an adaptive threshold.} 

To be more specific, as a \textit{warm-up} phase, for given integers $n_0$ and $T_0$, we first train a DGM with a conventional loss function using mini-batches with a fixed size of $n_0$ for $T_0$ updates. 
We use this non-truncated loss function to obtain an estimated parameter where the IM effect starts to appear.

After that, as the second phase, we apply the mini-batch increment and loss truncation strategies.  
At each update iteration $t$, we access a mini-batch $\mathcal{D}_t\subset\mathcal{D}^{\text{tr}}$ whose size is \textit{an exponential function of the iteration} $t$, i.e., $|\mathcal{D}_t|=n_0\gamma^{t-1}(=:n_t)$ for a constant $\gamma>1$.  
And instead of using all the samples included in $\mathcal{D}_t$ to calculate the loss function, we exploit the \textit{truncated loss function} which is formularized as:
\begin{eqnarray}
\label{eq:trun_loss}
\hat{L}(\theta,\tau_t;\mathcal{D}_t)=\frac{\sum_{\boldsymbol{x}\in\mathcal{D}_t}l(\theta;\boldsymbol{x})\cdot I(l(\theta;\boldsymbol{x})\le\tau_t)}{\sum_{\boldsymbol{x}'\in\mathcal{D}_t}I(l(\theta;\boldsymbol{x}')\le\tau_t)},
\end{eqnarray}
where $\tau_t>0$ is an adaptive threshold. 

Theoretically, we set $\tau_t$ to be the inlier risk, i.e., $\tau_t=L_i(\theta_{t-1})$, where $\theta_{t-1}$ is the estimated parameter at $(t-1)$-th update with an SGD-based optimizer.   
However, the computation of $\tau_t=L_i(\theta_{t-1})$ is infeasible in practice since we do not know the anomaly information of the training samples. 
Instead, we introduce the quantile as the value of $\tau_t$.
Specifically, for a pre-specified value $0<\rho<1$, we filter out the $100\times(1-\rho)$\% of the samples in the mini-batch that have the largest per-sample loss values.

\begin{figure}[t]
\centering
\includegraphics[width=0.4\columnwidth]{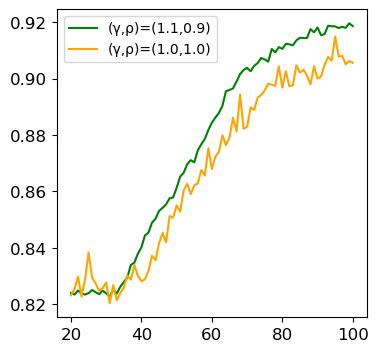} 
\includegraphics[width=0.4\columnwidth]{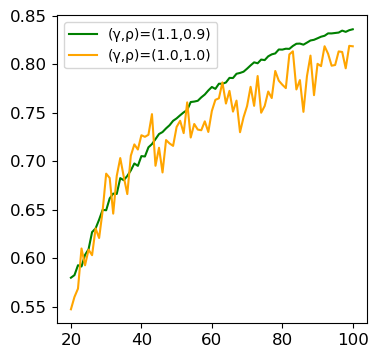}\\
\includegraphics[width=0.4\columnwidth]{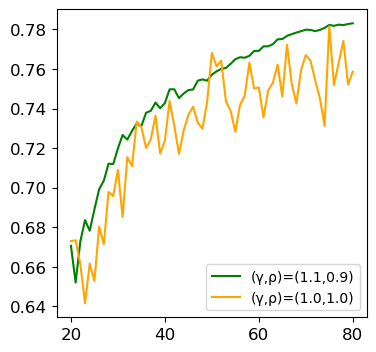} 
\includegraphics[width=0.4\columnwidth]{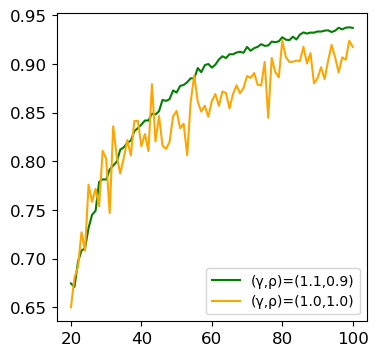} 
\caption{Outlier detection AUC values for DGMs with and without applying mini-batch increment and adaptive threshold, coloured as green and orange, respectively.
({\bf{Upper left} to \bf{clockwise}}) We analyze \texttt{Ionosphere, Letter, Vowels}, and \texttt{MagicGamma} datasets.
} 
\label{fig:altbi_effect}
\end{figure}

We conduct a simple experiment to empirically validate the effect of use of mini-batch increment and adaptive threshold. 
Two VAEs are trained in two scenarios—one with and the other without applying the two strategies. 
For the first scenario, we set $(T_0,n_0,\gamma,\rho)=(10,128,1.1,0.9)$. 
The outlier detection AUC results of the training data over four datasets are illustrated in Figure \ref{fig:altbi_effect}. 
We can clearly observe that the IM effect is more pronounced when using the two strategies, leading to superior performance in outlier detection. 
Additionally, using an increasing mini-batch size reduces fluctuation over updates, resulting in a more stable trained model. 
The theoretical properties of using adaptive mini-batch size and threshold will be discussed in the subsequent section.

\begin{remark}
Increasing the mini-batch size and using truncated loss function during training were first proposed in \citet{DBLP:conf/icml/XuSYQLSLJ21}. 
They utilized these techniques to develop enhanced classifiers in semi-supervised learning tasks. 
Our proposed method has its own contribution in that we find the close connection between the IM effect and these two techniques in the UOD regime and apply them to train DGMs with high outlier detection performance.
\end{remark}

\paragraph{Ensemble within a single model}
Recall that ODIM measures the degree of bi-modality in the per-sample loss distribution to find the optimal update. 
We empirically find that this heuristic approach often fails to identify the optimal update where the IM effect is maximized, sometimes even selecting an update where the IM effect does not appear. 
Additionally, ODIM employs an ensemble method to enhance performance using multiple under-fitted models, which increases computation time and resources.

To address this issue, we neither measure bi-modality nor use multiple models. 
Instead, we adopt the ensemble approach \textit{within a single model}.    
For given two integers $T_1,T_2$ with $T_1<T_2$, we take the average of per-sample loss values from $T_1+1$ to $T_2$ updates. 
That is, for a given input $\boldsymbol{x}$, we compute its outlier score as 
\begin{eqnarray}
\label{eq:ensemble}
{s}^{\text{ALTBI}}(\boldsymbol{x})=\frac{1}{T_2-T_1}\sum_{t=T_1+1}^{T_2}l(\theta_t;\boldsymbol{x}),    
\end{eqnarray}
where $\theta_t$ is the estimated parameter at the $t$-th update. 

It is obvious that using an ensemble approach with a single model for various updates leads to greater computational efficiency compared to considering multiple models. 
We demonstrate that this approach not only improves performance but also provides stability, as reported in the experimental section. 

We combine the above three techniques—1) mini-batch increment, 2) truncated loss, and 3) loss ensemble at various updates—to propose our method, which we term \textit{Adaptive Loss Truncation with Batch Increment (ALTBI)}. 
The pseudo algorithm of ALTBI is presented in Algorithm \ref{alg:algorithm}.

\paragraph{Choice of DGM framework}
There are numerous DGMs involved in likelihood maximization, such as VAE-based \citep{kingma2013auto,DBLP:journals/corr/BurdaGS15}, NF-based \citep{DBLP:conf/iclr/DinhSB17,DBLP:conf/nips/KingmaD18}, and score-based models \cite{DBLP:conf/nips/HoJA20,DBLP:conf/iclr/0011SKKEP21}. 
Among these, we decide to use two approaches: IWAE \citep{DBLP:journals/corr/BurdaGS15} and GLOW \citep{DBLP:conf/nips/KingmaD18}, which are widely used DGMs based on VAE and NF, respectively. 
Accordingly, their loss functions, $l(\boldsymbol{x};\theta)$, would be the ELBO-like upper bound and exact log-likelihood, respectively.   
Given that one of our method's key properties is computational efficiency, we do not consider score-based DGMs due to their large model sizes.


\begin{algorithm}[tb]
\caption{ALTBI\\ In practice, we set \\
$(n_0, \gamma, \rho, T_0, T_1, T_2) = (128, 1.03, 0.92,10,60,80)$.}
\label{alg:algorithm}
\textbf{Input}: Training data: $\mathcal{D}^{\text{tr}}=\left\{\boldsymbol{x}_j\right\}_{j=1}^n$,  parameters of a given DGM :$\theta$, initial mini-batch size: $n_0$, mini-batch increment: $\gamma$, quantile value: $\rho$, learning rate: $\eta$, three time steps: $T_0, T_1$, and $T_2$.

\begin{algorithmic}[1] 
\STATE Initialize $\theta_{0}$.\\
\STATE \textcolor{gray}{Phase 1: Warm-up}
\FOR{$(t=1$ \textbf{to} $T_0)$}
\STATE Draw a mini-batch with the fixed size of $n_0$, $\mathcal{D}_t=\{{\boldsymbol{x}}_{j}^{\text{mb}}\}_{i=1}^{n_0}$, from $\mathcal{D}^{\text{tr}}$. \\
Calculate the loss function: \\
$\hat{L}(\theta_0;\mathcal{D}_t)=\frac{1}{n_0} \sum_{j=1}^{n_0} l(\theta_0 ; \boldsymbol{x}_{j}^{\text{mb}})$. \\
Update $\theta_0$:\\ 
$\theta_{0} \leftarrow \theta_{0} - \eta\cdot\nabla_{\theta} \hat{L}(\theta_0;\mathcal{D}_t)$.
\ENDFOR
\STATE \textcolor{gray}{Phase 2: Enhancement of IM effect}
\FOR{$(t=1$ \textbf{to} $T_2 )$}
\STATE Draw a mini-batch with a size of $n_t = n_0 \gamma^{t-1}$, $\mathcal{D}_t=\{\boldsymbol{x}_{j}^{\text{mb}}\}_{j=1}^{n_t}$ from $\mathcal{D}^{\text{tr}}$. \\
Set the threshold $\tau_t=L_i(\theta_{t-1})$. \  \textcolor{gray}{// In practice, we choose $\tau_t$ as $ (100\times \rho)$-percentile of $\{l(\theta_{t-1};\boldsymbol{x}_j)\}_{j=1}^{n_t}$.} \\
Compute truncated loss $\hat{L}(\theta_{t-1}, \tau_t)$ as (\ref{eq:trun_loss})\\
Update $\theta_t$: \\
$\theta_{t} \leftarrow \theta_{t-1} - \eta \cdot\nabla_{\theta} \hat{L}(\theta_{t-1}, \tau_t)$.
\IF {$(t> T_1)$}
\STATE Incorporate the per-sample loss values to the final ALTBI scores as (\ref{eq:ensemble})
\ENDIF
\ENDFOR
\end{algorithmic}
\textbf{Output}: ALTBI scores of training data: $\left\{{s}^{\text{ALTBI}}(\boldsymbol{x}_j)\right\}_{j=1}^n$
\end{algorithm}

\section{Theoretical analysis}
\label{sec:theory}

In this section, we provide theoretical explanations to show that using the mini-batch size increment and truncated loss actually boosts the IM effect. 
We first state the rigorous definition of the IM effect below.

\begin{assumption}[IM effect]
\label{assump_im}
There exist $0<a_1<a_2<1$ and $a_3\in(0,1-a_2)$ such that for any parameter $\theta$ satisfying $L_{i}(\theta)\in[a_1,a_2]$, $L_{o}(\theta)-L_{i}(\theta)\ge a_3$.
\end{assumption}

Assumption \ref{assump_im} refers to the property that, when a given DGM is trained for a while, there is a gap in risk values between inliers and outliers. 
A couple of additional yet reasonable assumptions about the gradient are required, which are almost the same as those in \citet{DBLP:conf/icml/XuSYQLSLJ21}. 

\begin{assumption}[Bounded and smooth gradient]
\label{assump_grad}
Denote the gradients of $l(\theta;\boldsymbol{x})$ and $L_i(\theta)$ as $\nabla_{\theta} l(\theta;\boldsymbol{x})$ and $\nabla_{\theta} L_i(\theta)$, respectively. Then the followings conditions are satisfied:
\begin{enumerate}

    \item[(i)] For any $\boldsymbol{x}\in\mathcal{X}$ and $\theta\in\Theta$, there exists a constant $G > 0$, such that
    \begin{align*}
    \|\nabla_{\theta} l(\theta;\boldsymbol{x})\| \leq G.
    \end{align*}
    \item[(ii)] $L_i(\theta)$ is smooth with a $L$-Lipschitz continuous gradient, i.e., there exists a constant $L > 0$ such that
    \begin{align*}
    \|\nabla_{\theta} L_i(\theta) - \nabla_{\theta} L_i(\theta')\| \leq L \|\theta - \theta'\|, \quad \forall \theta, \theta' \in \Theta,
    \end{align*}
    \item[(iii)] There exists $\mu > 0$ such that for any $\theta\in\Theta$,
    \begin{align*}
    2\mu (L_i(\theta) -L_i(\theta_*))=2\mu L_i(\theta) \leq \|\nabla_{\theta} L_i(\theta)\|^2, 
    \end{align*}
    where $\theta_*$ is the minimizer of $L_i(\theta)$. 

    \end{enumerate}
\end{assumption}
The first and second assumptions in Assumption \ref{assump_grad} refer to the properties that the loss function and inlier risk are smooth. 
The last assumption is known as the Polyak-\L{}ojasiewicz condition \citep{polyak1964gradient}, which is widely considered in the literature related to SGD with deep learning (\citet{DBLP:conf/nips/Yuan0JY19} and references therein). 

We finally introduce a technical assumption about the loss distribution. 
We note that this condition is quite weak and can be satisfied in general situations.  
\begin{assumption}[Loss distribution]
There is a constant $0<c<1$ such that, for any $\theta$, the following inequality holds: 
$$\bigl[E_{P_i}\sqrt{l(\theta;X)}\bigr]^{2}\le (1-c)L_i(\theta).$$     
\end{assumption}

Then we have the following proposition, which asserts that if we apply the mini-batch increment and threshold to truncate the loss function, the ratio of outliers included in the truncated loss becomes small.
The proof of Proposition \ref{prop:1} is provided in the Appendix A. 

\setcounter{theorem}{0}
\begin{proposition}
\label{prop:1}
At the $t$-th update, we suppose that the current parameter $\theta_{t-1}$ satisfies $a_1\le L_i(\theta_{t-1})\le a_2\gamma^{-(t-1)}$. 
For a mini-batch $\mathcal{D}_t$, we denote the inlier set which is included in the truncated loss as $\mathcal{A}_t^{\tau}$. 
Similarly, we can define $\mathcal{B}_t^{\tau}$ for outliers. 
Then, under Assumptions 1 to 3 and for a given $\delta>0$, there exists positive constants $c_1$ and $c_2$ not depending on $t$ such that the following two inequalities holds:
\begin{align*}
|\mathcal{A}_t^{\tau}|\ge c_1 n_t\quad {and}\quad|\mathcal{B}_t^{\tau}|\le c_2 n_0,
\end{align*}
with a probability at least $1-\delta$.
\end{proposition}

Considering $n_t=n_0\gamma^{t-1}$, Proposition \ref{prop:1} indicates that at the $t$-th update, the ratio of outliers included in the truncated loss cannot exceed $(c_2/c_1)\cdot\gamma^{-(t-1)}$, which decreases toward zero as the update step $t$ increases as long as the IM effect persists at each update. 
Therefore, our proposed method gradually refines samples in the loss function, leading to the clearer IM effect. 

\begin{figure}[t]
\centering
\includegraphics[width=0.47\columnwidth]{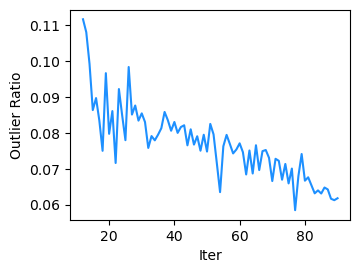} 
\includegraphics[width=0.47\columnwidth]{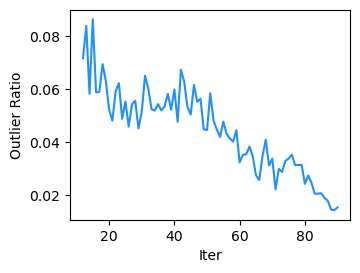}
\caption{Trace plot of outlier ratio in truncated samples over various iterations. 
We visualize two datasets: ({\bf{Left}}) \texttt{Cardio} and ({\bf{Right}}) \texttt{Shuttle}.
}
\label{fig:altbi_outlier}
\end{figure}

We visualize whether the ratio of outliers actually decreases as the updates proceed. The same learning framework and hyperparameter settings from Figure \ref{fig:outlier_ratio} are considered, and two datasets are analyzed: \texttt{Cardio} and \texttt{Shuttle}. 
Figure \ref{fig:altbi_outlier} shows that the outlier ratio in the truncated loss function tends to decrease over updates, providing empirical evidence for Proposition \ref{prop:1}.

We note that Proposition \ref{prop:1} holds provided that the inlier risk is sufficiently small, i.e., smaller than $a_2\gamma^{-(t-1)}$. 
Next theoretical result deals with the guarantee that the inlier risk indeed decreases over updates with high probability. 
The proof is provided in Appendix A. 

\begin{proposition}
\label{prop:2} 

At the $t$-th update, we suppose that all the assumptions considered in Proposition \ref{prop:1} hold. 
Then for a given $\delta>0$, there exists a learning rate $\eta>0$ such that $L_i(\theta_{t})\le a_2\gamma^{-t}$ with a probability at least $1-\delta$.
\end{proposition}

The above proposition implies that if the previously estimated DGM satisfies the IM effect with a small inlier risk, then the subsequent estimated DGM has a smaller inlier risk with a factor of $\gamma$.  

Suppose that the IM effect starts to be observed with the estimated parameter after warm-up step, i.e., $a_1\le L_i(\theta_0)\le a_2$. 
Then, Proposition \ref{prop:2} suggests that with a carefully chosen learning rate, ideally, we can observe an enhanced IM effect up to $\lfloor (\log(a_1 / a_2)) / (\log(1/\gamma)) \rfloor$ consecutive updates.

\section{Experiments}
\label{sec:exp}
We validate the superiority of our proposed method by analyzing an extensive set of 57 datasets, including image, text, and tabular data. We prove that ALTBI is the state-of-the-art solution for various types of data with its high computational efficiency compared to other recent methods. In each experiments, we report the averaged results based on three trials with random parameter initializations. We use the \texttt{Pytorch} framework to run our algorithm using a single NVIDIA TITAN XP GPU.

\paragraph{Dataset description} 
We analyze all 57 outlier detection benchmark datasets from \texttt{ADBench} \citep{han2022adbench}, including tabular, image, and text data.  
And as done in \citet{kim2024odimoutlierdetectionlikelihood}, we conduct the min-max scaling to pre-process each dataset. 
We first consider 46 widely used tabular datasets that cover various application domains, including healthcare, finance, and astronautics. 
Additionally, we include five benchmark datasets commonly used in the field of natural language processing (NLP): \texttt{20news}, \texttt{Agnews}, \texttt{Amazon},  \texttt{IMDB}, and \texttt{Yelp}. 
For these datasets, we utilize embedding features of these datasets via BERT \citep{devlin-etal-2019-bert} or RoBERTa \citep{DBLP:journals/corr/abs-1907-11692}, both publicly accessible in \texttt{ADBench}. 
We finally analyze six image datasets: \texttt{CIFAR10}, \texttt{MNIST-C}, \texttt{MVTec-AD}, \texttt{SVHN}, \texttt{MNIST}, and \texttt{FMNIST}. 
These datasets are analyzed using the embedding features extracted by the ViT \citep{DBLP:conf/iclr/DosovitskiyB0WZ21}, also available in \texttt{ADBench}. 
The detailed information about all the datasets is provided in the Appendix B and \citet{han2022adbench}.  

\paragraph{Baseline} 
We mainly compare our method with ODIM \citep{kim2024odimoutlierdetectionlikelihood}, and also consider other baselines compared in the study.  
These methods contain traditional machine-learning-based approaches whose implementations are provided in \texttt{ADBench}, including kNN \citep{ramaswamy2000efficient}, LOF \citep{breunig2000lof}, OCSVM \citep{ocsvm}, CBLOF \citep{he2003discovering}, PCA \citep{shyu2003novel}, FeatureBagging \citep{lazarevic2005feature}, IForest \citep{liu2008isolation}, MCD \citep{fauconnier2009outliers}, HBOS \citep{goldstein2012histogram}, LODA \citep{pevny2016loda}, COPOD \citep{li2020copod}, and ECOD \citep{li2022ecod}. 

And we also consider two deep learning-based UOD methods, DAGMM \citep{dagmm} and DeepSVDD \citep{deepsvdd}, both of which can be implemented through \texttt{ADBench}. 
Additionally, we evaluate our method against more recent deep learning approaches beyond \texttt{ADBench}, such as DROCC \citep{DBLP:conf/icml/GoyalRJS020}, ICL \citep{icl}, GOAD \citep{goad}, DTE \citep{DBLP:journals/corr/abs-2305-18593}, and ODIM \citep{kim2024odimoutlierdetectionlikelihood}.

\paragraph{Implementation details} 
As mentioned previously, we use two likelihood-based DGM frameworks: 1) IWAE \citep{DBLP:journals/corr/BurdaGS15}, an ELBO-based model and 2) GLOW \citep{DBLP:conf/iclr/NalisnickMTGL19}, an NF-based model. 
IWAE uses multiple latent vectors to make the objective function tighter than the standard ELBO. 
We use the same DNN architecture for IWAE as in \citet{kim2024odimoutlierdetectionlikelihood}, but set the number of the latent samples $K$ to two. 
Detailed descriptions of the architectures and loss functions are presented in Appendix B. 
And GLOW \citep{DBLP:conf/nips/KingmaD18} introduces invertible $1\times 1$ convolution filters to create normalizing flows with high complexity. 
The architecture considered in \citet{DBLP:conf/iclr/NalisnickMTGL19} is used, and we reshape each dataset into a squared form to apply this architecture. 

For the optimizer, we use Adam \citep{kingma2014adam} with a learning rate of $1e-3$. 
Throughout our experimental analysis, we fix the hyperparameters, necessary for our proposed method--$(n_0,\gamma,\rho,T_0,T_1,T_2)$--to (128,1.03,0.92,10,60,80), unless stated otherwise.  
Performance results for other hyperparameter values are provided in the ablation studies.

\begin{figure}[t]
\centering
\includegraphics[width=1\columnwidth]{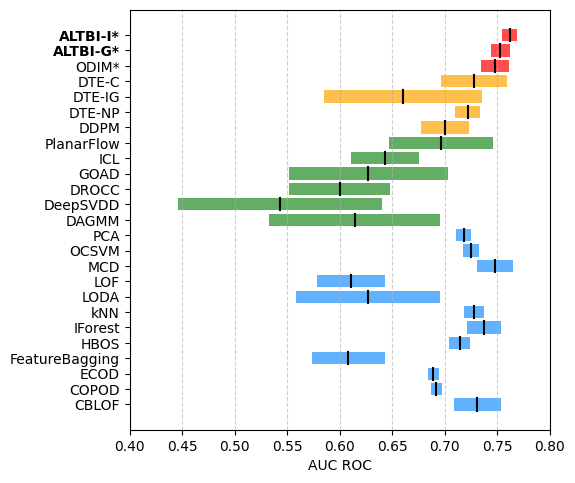} 
\caption{Averaged AUC results, including means and standard deviations, across 57 datasets from ADBench over three different implementations.
We mark an asterisk (*) next to methods for our own implementations. 
Color scheme: red (IM-based), orange (diffusion-based), green (deep-learning-based), blue (machine-learning-based).}
\label{fig:performace_results}
\end{figure}
\begin{figure*}[t]
\centering
\includegraphics[width=0.16\textwidth]{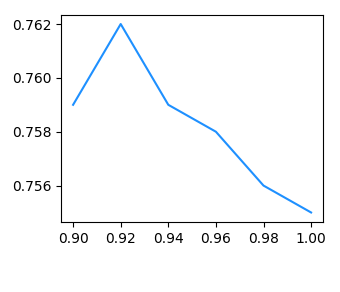}
\includegraphics[width=0.16\textwidth]{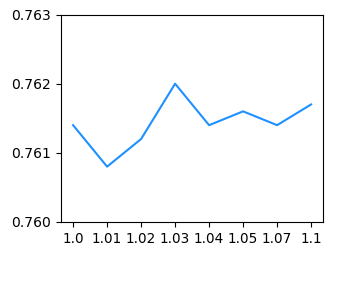}
\includegraphics[width=0.16\textwidth]{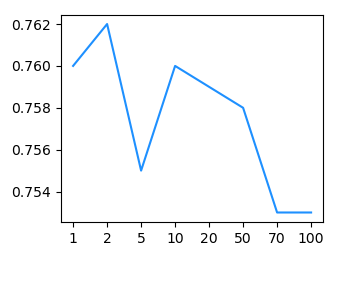}
\includegraphics[width=0.16\textwidth]{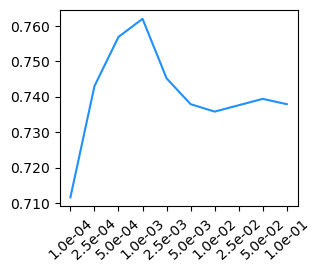}
\includegraphics[width=0.16\textwidth]{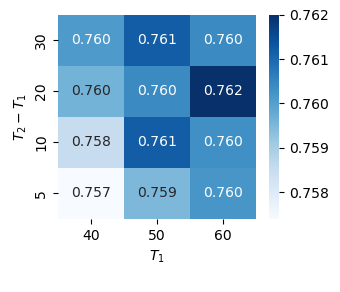}
\includegraphics[width=0.16\textwidth]{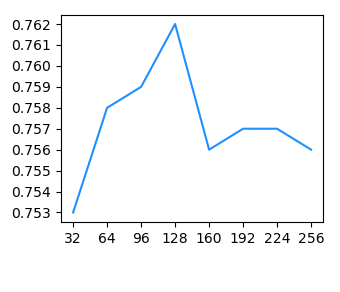}
\caption{(\textbf{From left to right}) 1) AUC scores with various values of $\rho$. 2) AUC scores with various values of $\gamma$. 3) AUC scores with various values of K in IWAE. 4) AUC scores with various values of learning rate. 5) Heatmap of AUC scores for ensembling with various values of $T_1$ and $T_1 - T_2$. 6) AUC scores with various values of $n_0$.} 
\label{fig:ablation_study}
\end{figure*}
\subsection{Performance results}
We evaluate the outlier detection performance of ALTBI with IWAE (ALTBI-I) and GLOW (ALTBI-G) in comparison with other baselines. 
For each dataset, the mean and standard deviation of outlier detection AUC and PRAUC over three different implementations are measured. 
We report the averaged means and standard deviations of AUC across all datasets in Figure \ref{fig:performace_results}. 
Detailed results for each dataset, including PRAUC, are summarized in Appendix B. 
We acknowledge that we implemented ALTBI and ODIM ourselves, while all other baseline results are referenced from the Appendix in \citet{DBLP:journals/corr/abs-2305-18593}.

Figure \ref{fig:performace_results} shows that the scores of ALTBI-I achieves the best performance, followed by ALTBI-G, both outperforming all other baselines. 
Considering the computational efficiency of IWAE compared to GLOW, ALTBI-I could be a more favorable method as a UOD solver. 
Additionally, our method showcases smaller standard deviations compared to other baselines. 
These results indicate that ALTBI has superior and stable performance across various data types in empirical experiments, again highlighting ALTBI as a off-the-shelf method for UOD.

\subsection{Ablation studies}
We perform further experiments to explore the impact of hyperparameter choices on ALTBI's performance across the \texttt{ADBench} datasets and the results are presented in Figure \ref{fig:ablation_study}. 
Detailed results can be found in the Appendix B.

\noindent
\textcircled{1} The performance improves as the truncation percentage increases up to 8$\%$, but it starts to decline afterward.\\ 
\textcircled{2} Increasing the mini-batch size by a factor of $\gamma=1.03$ at each update leads to optimal performance.\\
\textcircled{3} Performance is best when $K=2$ in IWAE, and it tends to decline as $K$ increases further.\\
\textcircled{4} The learning rate increases up to $1e-3$, resulting in continuous performance improvement, but after that, performance decreases and stabilizes.\\
\textcircled{5} Finding appropriate values of $T_1$ and $T_2$ for ensembling within a DGM single model affects ALTBI's performance but the impact is not significant.\\
\textcircled{6} Increasing $n_0$ enhances ALTBI's performance up to a value of 128, after which the performance begins to decline and stabilize.\\
\textcircled{7} Additionally, we empirically find that ALTBI achieve near state-of-the-art performance in solving SSOD tasks as well. 
We provide verification of this in Appendix B.

\subsection{Further discussions: Robustness of ALTBI in DP}
\begin{figure}[t]
\centering
\includegraphics[width=0.5\columnwidth]{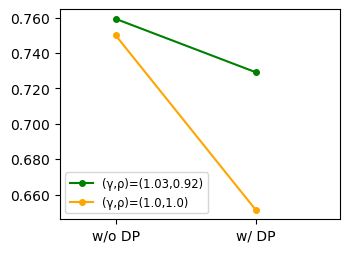} 
\caption{The impact of mini-batch increment and loss truncation when applying an DP-SGD algorithm.}
\label{fig:dp_altbi}
\end{figure}

A representative method to ensure that a given algorithm satisfies differential privacy (DP) is by training with the DP-SGD algorithm \citep{abadi2016deep} instead of conventional SGDs. 
This involves clipping the gradient norm for each per-sample loss and adding Gaussian noise.
For a given loss function $\tilde{l}(\theta;\boldsymbol{x})$, this operation can be formularized as 
\begin{align*}
\text{Clip}\bigl(\nabla_{\theta}\tilde{l}(\theta,\boldsymbol{x});C\bigr)+\mathcal{N}(0,\sigma^2C^2 I),
\end{align*}
where $C>0$ is a clipping constant and $\sigma>0$ controls the noise amount. 

We note that ALTBI utilizes $\tilde{l}(\theta;\boldsymbol{x})=l(\theta;\boldsymbol{x})I(l(\theta;\boldsymbol{x})\le \tau)$. 
When a sample is filtered out from the truncated loss, its gradient is already clipped to have a norm of zero. 
Since outliers are mostly excluded by the truncated loss, leading to their gradients being clipped to zero, we can infer that incorporating DP-SGD into ALTBI preserves the inliers' information relative to outliers, making ALTBI inherently robust when implementing DP-SGD.

To validate our claim, we conduct an additional experiment by analyzing 20 tabular datasets. 
We consider two versions of ALTBI: one that applies mini-batch increment and truncated loss, i.e., $(\gamma,\rho)=(1.03,0.92)$, and one that does not,i.e., $(\gamma,\rho)=(1.0,1.0)$. 
As a measure of DP, we adopt $(\epsilon,\delta)$-DP, and with a fixed $\delta=1e-5$, we train them using a DP-SGD algorithm until the cumulative privacy budget $\epsilon\le 10$ holds. 
Then we compare the averaged outlier detection AUC values in Figure \ref{fig:dp_altbi}. 
The modified ALTBI for DP and detailed results are provided in Appendix B. 

Figure \ref{fig:dp_altbi} shows that increasing mini-batch sizes and using the truncated loss function yields more robust performance when applying the DP algorithm, indicating that combining ALTBI and DP has a synergistic effect.

\section{Concluding remarks}
\label{sec:conclusion}
In this study, we developed a novel UOD method called ALTBI, which maximally exploits the IM effect. 
By introducing two key techniques--gradually increasing the mini-batch size and adopting an adaptive threshold to truncate the loss function--ALTBI demonstrated superior and stable outlier detection performance across various datasets while maintaining computational efficiency. 
Extensive experiments validated ALTBI's state-of-the-art results, making it a robust and effective solution for UOD.
Several studies have extended outlier detection tasks to scenarios where a few outliers with known outlier information are accessible \citep{deepsad,kim2024odimoutlierdetectionlikelihood}. 
Applying our method to the more complex case where some labeled outliers are wrongly annotated would be an interesting direction for future work.

\bibliography{references}
\bibliographystyle{icml2023}

\newpage
\appendix
\onecolumn

\section{Proof of Theoretical results}
Our theoretical results and their proofs are similar to Theorem 1 in \citet{DBLP:conf/icml/XuSYQLSLJ21}. 
Before starting to prove Proposition 1\&2, we first state three lemmas that are used throughout our proofs. 

\begin{lemma}
\label{lem:1}
If the IM assumption is satisfied, there exists $a_4>0$ such that  the following inequality holds:
\bea
\label{im}
P_o\left(l(\theta;X)\le L_i(\theta) \right)\le a_4\cdot L_i(\theta).
\eea
\end{lemma}

\noindent
\textit{proof)}
We have the following inequalities:
\begin{eqnarray*}
P_o(l(\theta;X)\le L_i(\theta))&=&P_o(l(\theta;X)-L_o(\theta)\le -(L_o(\theta)-L_i(\theta)) \\
&\le& P_o(|l(\theta;X)-L_o(\theta)|\ge (L_o(\theta)-L_i(\theta))\\
&\overset{\text{(Markov's Ineq.)}}{\le}& \frac{\E_o|l(\theta;X)-L_o(\theta)|}{L_o(\theta)-L_i(\theta)}\\
&\le& \frac{\E_o|l(\theta;X)|+L_o(\theta)}{L_o(\theta)-L_i(\theta)}
\le \frac{2L_o(\theta)}{a_3}.
\end{eqnarray*}
Since $a_1\le L_i(\theta)$, we have $L_o(\theta)\le 1\le L_i(\theta)/a_1$. 
Therefore, we have
\begin{equation*}
P_o(l(\theta;X)\le L_i(\theta))\le \frac{2L_o(\theta)}{a_3}
\le \frac{2}{a_1 a_3}L_i(\theta),
\end{equation*}
and the proof is completed with $a_4=2/(a_1 a_3)$. \qed

\begin{lemma}[Conditional version of Theorem 3.6 in \citet{chung2006concentration}]
\label{lem:2}
Suppose that $Y_i$ for $i\in[n]$ are random variables satisfying $Y_i\le M$ and $\mathcal{H}$ is a given $\sigma$-algebra. 
Let assume that the conditional expectations $\E(Y_i|\mathcal{H})$s are independent. 
Let $Y=\sum_{i=1}^n Y_i$ and $||Y||_{\mathcal{H}}=\sqrt{\sum_{i=1}^n \E(Y_i^2|\mathcal{H})}$. 
Then, for any $\lambda>0$, we have
\begin{align*}
P\left(Y\ge \E(Y|\mathcal{H})+\lambda\Big|\mathcal{H}\right) \le \exp{\left(-\frac{\lambda^2}{2(||Y||_{\mathcal{H}}^2+M\lambda/3)}\right)} \text{ a.s.}.
\end{align*}
\end{lemma}

\begin{lemma}[Conditional version of Lemma 4 in \citet{ghadimi2016mini}]
\label{lem:3}
Suppose that $Y_i$s, $i\in[n]$, are random variables with mean zero and $\mathcal{H}$ is a given $\sigma$-algebra. 
Let us assume that there exist positive values $\sigma_i^2>0$, $i\in[n]$, such that $E(Y_i)^2\le \sigma_i^2$. 
We also assume that the conditional expectations $\E(Y_i|\mathcal{H})$s are independent. 
Then for any $\lambda>0$, the following holds:
\begin{align*}
P\left(  \left\| \sum_{i=1}^n Y_i^2\right\| \ge \lambda\sum_{i=1}^n \sigma_i^2 \Big| \mathcal{H} \right)\le \frac{1}{\lambda} \text{ a.s.}.
\end{align*}
\end{lemma}

\subsection{Proof of Proposition 4.1}
\noindent
\textreferencemark We prove this proposition with a probability of $1-4\delta$. 
Transforming $1-4\delta$ to $1-\delta$ is trivial by substituting $\delta$ with $\delta/4$.\\

\noindent
Before we carry out our analysis, we define a few important constants
\begin{align*}
C_1 &= \sqrt{\frac{\log(2/\delta)}{2c^2(1-\alpha)n_0}},\\
C_2 &= \max \left( \sqrt{\frac{\log(2/\delta)}{2(1-\alpha))^2 n_0}}, \sqrt{\frac{\log(2/\delta)}{2\alpha^2 n_0}} \right),\\
C_3 &= c(1-\alpha)( 1-C_1)(1-C_2),\\
C_4 &= \left(4\alpha(1+C_2) a_2 a_4 +\frac{1}{3}\log(1/\delta)\right),
\end{align*}
where $a_4$ is the same constant as in Lemma \ref{lem:1}.
For the mini-batch $\mathcal{D}_t=\{X_1,\ldots,X_{n_t}\}$ with $n_t = n_0 \gamma^{t-1}$ training examples sampled from $\mathcal{D}$, we divide it into two sets for the analysis use only, i.e. set $\mathcal{A}_t$ that includes examples sampled from $\mathcal{P}_i$ and set $\mathcal{B}_t$ that includes examples sampled from $\mathcal{P}_o$. We furthermore denote by $\mathcal{A}_t^\tau$ and $\mathcal{B}_t^\tau$ the subset of examples in $\mathcal{A}_t$ and $\mathcal{B}_t$ whose loss is smaller than the given threshold $\tau_t$, i.e.
\begin{align*}
\mathcal{A}_t^\tau = \{X \in \mathcal{A}_t : l(\theta_{t-1}; X) \leq \tau_t\},
\end{align*}
\begin{align*}
\mathcal{B}_t^\tau = \{X \in \mathcal{B}_t : l(\theta_{t-1}; X) \leq \tau_t\},
\end{align*}
where $\tau_t = L_i(\theta_{t-1})$ and $\theta_{t-1}$ is the currently estimated parameter at the $(t-1)$-th update. 
We assume that $L_i(\theta_{t-1})\le a_2\gamma^{-(t-1)}$. 
Evidently, the samples used for computing $\mathbf{g}_t$ are the union of $\mathcal{A}_t^\tau$ and $\mathcal{B}_t^\tau$.
The following result bounds the size of $\mathcal{A}_t^\tau$ and $\mathcal{B}_t^\tau$. With a probability $1-4\delta$. 
We have 
\begin{align*}
    |\mathcal{A}_t^\tau| \ge C_3 n_t=C_3 n_0\gamma^{t-1} ,\ \  |\mathcal{B}_t^\tau | \le C_4 n_0,
\end{align*}
where $C_3$ and $C_4$ are defined above.

\paragraph{(i) Lower bound of $\mathcal{A}_t^{\tau}$}
By Hoeffding's inequality, for any $t>0,$ we have
\begin{align*}
P_*(||\mathcal{A}_t|-(1-\alpha)n_t|\ge -t)\ge 
P_*(||\mathcal{A}_t|-(1-\alpha)n_t|\le t)\ge 1-2\exp(-2t^2/n_t).
\end{align*}
By substituting $t=\sqrt{\frac{n_t\log(2/\delta)}{2}}$ with $\delta>0$, we have
\begin{align}
\label{ineq1}
|\mathcal{A}_t|\ge (1-\alpha)n_t \left(1-\sqrt{\frac{\log(2/\delta)}{2(1-\alpha)^2n_t}}\right)\ge (1-\alpha)(1-C_2)n_t,
\end{align}
with a probability at least $1-\delta$. 

\noindent
By using the general Markov inequality,  
\begin{align*}
P_i(l(\theta_{t-1} ; X) \le \tau_t) = 1- P_i(l(\theta_{t-1} ; X) \ge \tau_t) \ge 1- \frac{\E_{P_i}[\sqrt{l(\theta_{t-1};X)}]}{\sqrt{L_i(\theta_{t-1})}}\ge c, 
\end{align*}
where the last inequality holds due to Jensen's inequality. 
Let $\mathcal{F}_t:=\mathcal{F}\left(I(X_j\in\mathcal{X}_i),j\in[n_t]\right)$. 
We apply the conditional Hoeffding's inequality to achieve the following: for any $t>0$,
\begin{align*}
P_*\Bigl(|\mathcal{A}_t^{\tau}|-c|\mathcal{A}_t|\ge -t \Big|\mathcal{F}_t\Bigr)&\ge
P\Bigl(\big| |\mathcal{A}_t^{\tau}|-\E (|\mathcal{A}_t^{\tau}|) \big|\le t \Big|\mathcal{F}_t\Bigr)\\
&\ge 1-2\exp\bigl( -\frac{2t^2}{|\mathcal{A}_t|}\bigr)\text{ a.s.}
\end{align*}
With $t=\sqrt{\frac{|\mathcal{A}_t|}{2}\log(2/\delta)}$, 
\begin{align*}
P_*\biggl(|\mathcal{A}_t^{\tau}|\ge c|\mathcal{A}_t| \biggl( 1-\sqrt{\frac{\log(2/\delta)}{2c^2 |\mathcal{A}_t|}} \biggr)  \Big|\mathcal{F}_t\biggr)\ge 1-\delta\text{ a.s.},
\end{align*}
and hence
\begin{align}
\label{ineq3}
P_*\biggl(|\mathcal{A}_t^{\tau}|\ge c|\mathcal{A}_t| \biggl( 1-\sqrt{\frac{\log(2/\delta)}{2c^2 |\mathcal{A}_t|}} \biggr)\biggr)\ge 1-\delta.
\end{align}
Combining (\ref{ineq1}) and (\ref{ineq3}), we have 
\begin{align}
\label{ineq9}
|\mathcal{A}_t^{\tau}|&\ge c|\mathcal{A}_t| \biggl( 1-\sqrt{\frac{\log(2/\delta)}{2c^2 |\mathcal{A}_t|}} \biggr)\nonumber\\
&\ge c(1-\alpha)(1-C_2)n_t  \biggl( 1-\sqrt{\frac{\log(2/\delta)}{2c^2 (1-\alpha)(1-C_2)n_t}} \biggr) \nonumber\\
&\ge c(1-\alpha)(1-C_2)( 1-C_1)n_t \nonumber\\
&= C_3 n_t,
\end{align}
with a probability at least $1-2\delta$.

\paragraph{(ii) Upper bound of $\mathcal{B}_t^{\tau}$}
And by Hoeffiding's inequality, we also have
\begin{align*}
P_*(|\mathcal{B}_t|-\alpha n_t\le t)\ge 
P_*(||\mathcal{B}_t|-\alpha n_t|\ge t)\ge 1-2\exp(-2t^2/n_t).
\end{align*}
By substituting $t=\sqrt{\frac{n_t\log(2/\delta)}{2}}$, we have
\begin{align}
\label{ineq2}
|\mathcal{B}_t|\le \alpha n_t \left( 1+\sqrt{\frac{\log(2/\delta)}{2\alpha^2 n_t}} \right)\le \alpha(1+C_2) n_t,
\end{align}
with a probability at least $1-\delta.$
Also, by using Lemma \ref{lem:1}, the following inequalities hold:
\begin{align}
\label{ineq4}
P_o (l(\theta_{t-1}; X) \leq \tau_t) &= P_o (l(\theta_{t-1}; X) \leq L_i(\theta_{t-1}))
\stackrel{(Lem. \ref{lem:1})}{\leq} a_4\cdot L_i(\theta_{t-1}) 
\le a_2 a_4 \gamma^{-(t-1)}.
\end{align}

\noindent
Let $\mathcal{G}_t:=\mathcal{F}\left(I(X_j\in\mathcal{X}_o),j\in[n_t]\right)$. 
We can bound the expectation of $|\mathcal{B}_t^\tau|$ given $\mathcal{G}_t$, i.e.,
\begin{align}
\label{ineq7}
\E_*\Bigl[|\mathcal{B}_t^{\tau}|\Big| \mathcal{G}_t\Bigr] &=
\E_*\left[\sum_{X\in\mathcal{B}_t}I\bigl(l(\theta_{t-1};X)\le \tau_t \bigr)\Big| \mathcal{G}_t\right]\nonumber \\
&{\le} |\mathcal{B}_t|P_o\bigl( l(\theta_{t-1};X)\le\tau_t \bigr)
\stackrel{(\ref{ineq4})}{\le} |\mathcal{B}_t|a_2 a_4 \gamma^{-(t-1)} \text{ a.s.}.
\end{align}
Also, by using Lemma \ref{lem:2}, for any $\lambda>0$, the following inequality holds:
\begin{align*}
P_*\left( |\mathcal{B}_t^{\tau}|\le \E_*\left[|\mathcal{B}_t^{\tau}| \big|\mathcal{G}_t\right] +\lambda\Big|\mathcal{G}_t \right)\ge 
1-\exp{\left( -\frac{\lambda^2}{2 { \E_*\Bigl[|\mathcal{B}_t^{\tau}|\Big| \mathcal{G}_t\Bigr]+\lambda/3} }\right)}
\text{ a.s.}
\end{align*}
For a given $\delta>0$, by substituting $\lambda=\frac{1}{3}\log(1/\delta)+\sqrt{\frac{1}{9}\log^2(1/\delta)+2\log(1/\delta)\E_*\bigl[|\mathcal{B}_t^{\tau}|\big| \mathcal{G}_t\bigr]}$, we have
\begin{align*}
 P_*\left( |\mathcal{B}_t^{\tau}|\le \E_*\left[|\mathcal{B}_t^{\tau}| \big|\mathcal{G}_t\right] +\frac{1}{3}\log(1/\delta)+\sqrt{\frac{1}{9}\log^2(1/\delta)+2\log(1/\delta)\E_*\bigl[|\mathcal{B}_t^{\tau}|\big| \mathcal{G}_t\bigr]}\Big|\mathcal{G}_t \right)\ge 
1-\delta \text{ a.s.},
\end{align*}
and hence
\begin{align}
\label{ineq5}
P_*\left( |\mathcal{B}_t^{\tau}|\le \E_*\left[|\mathcal{B}_t^{\tau}| \big|\mathcal{G}_t\right] +\frac{1}{3}\log(1/\delta)+\sqrt{\frac{1}{9}\log^2(1/\delta)+2\log(1/\delta)\E_*\bigl[|\mathcal{B}_t^{\tau}|\big| \mathcal{G}_t\bigr]} \right)\ge 
1-\delta.
\end{align}
We combine (\ref{ineq2}), (\ref{ineq7}), and (\ref{ineq5}) to achieve the inequality below:
\begin{align}
\label{ineq8}
|\mathcal{B}_t^{\tau}| &\le \E_*\left[|\mathcal{B}_t^{\tau}| \big|\mathcal{G}_t\right] +\frac{1}{3}\log(1/\delta)+\sqrt{\frac{1}{9}\log^2(1/\delta)+2\log(1/\delta)\E_*\bigl[|\mathcal{B}_t^{\tau}|\big| \mathcal{G}_t\bigr]} \nonumber\\
&\le 4\E_*\left[|\mathcal{B}_t^{\tau}| \big|\mathcal{G}_t\right]+\frac{2}{3}\log(1/\delta)\nonumber\\
&\le 4|\mathcal{B}_t|a_2 a_4 \gamma^{-(t-1)}+\frac{2}{3}\log(1/\delta)\nonumber\\
&\le 4\alpha(1+C_2) n_0 \gamma^{t-1} a_2 a_4 \gamma^{-(t-1)}+\frac{2}{3}\log(1/\delta)\nonumber\\
&= 4\alpha(1+C_2) n_0 a_2 a_4 +\frac{2}{3}\log(1/\delta)\nonumber \\
&\le \left(4\alpha(1+C_2) a_2 a_4 +\frac{1}{3}\log(1/\delta)\right) n_0 =C_4 n_0.
\end{align}
with a probability at least $1-2\delta$.

\noindent
Therefore, with (\ref{ineq9}) and (\ref{ineq8}), the proof is completed with $c_1=C_3$ and $c_2=C_4$. \qed

\subsection{Proof of Proposition 4.2}
\textreferencemark Similar to the proof of Proposition 1, we prove Proposition 2 with a probability of $1-5\delta$. 
Transforming $1-5\delta$ to $1-\delta$ can be done by using $\delta/5$ instead of $\delta$. \\

\noindent
We will demonstrate that, with high probability, $L(\theta_{t}) \le a_2\gamma^{-t}$. 
Let $\mathbf{g}_t=\nabla_{\theta}\hat{L}(\theta_{t-1},\tau_t)$.
To this end, using the notation of $\mathcal{A}_t^\tau$ and $\mathcal{B}_t^\tau$, we can rewrite $\mathbf{g}_t$ as 
\begin{align*}
    \mathbf{g}_t = (1-b_t)\mathbf{g}_t^a + b_t \mathbf{g}_t^b, 
\end{align*}
where $\mathbf{g}_t^a = \frac{1}{|\mathcal{A}_t^\tau|}\sum_{X\in\mathcal{A}_t^\tau}\nabla l(\theta_{t-1};X)$,\ \ $\mathbf{g}_t^b = \frac{1}{|\mathcal{B}_t^\tau|}\sum_{X\in\mathcal{B}_t^\tau}\nabla l(\theta_{t-1};X)$, and $b_t$ is the proportion of samples from $\mathcal{B}_t^\tau$ that 
\begin{align*}
    b_t = \frac{|\mathcal{B}_t^\tau|}{|\mathcal{A}_t^\tau|+|\mathcal{B}_t^\tau|} \le \frac{|\mathcal{B}_t^\tau|}{|\mathcal{A}_t^\tau|} \le \frac{c_2}{1+c_1\gamma^{t-1}}<\frac{c_2}{c_1}\gamma^{-(t-1)},
\end{align*}
with a probability of at least $1-4\delta$ by Proposition 1.
Following the classical analysis of non-convex optimization, since $L(\theta)$ is $L$-smooth by Assumption 2 (ii), we have 
\begin{align}
\label{ineq12}
L_i(\theta_{t}) - L_i(\theta_{t-1}) &\overset{(a)}{\leq} \langle \nabla L_i(\theta_{t-1}), \theta_{t} - \theta_{t-1} \rangle + \frac{L}{2} \|\theta_{t} - \theta_{t-1}\|^2 \nonumber\\
&\overset{(b)}{=} \frac{\eta}{2} \|\nabla L_i(\theta_{t-1}) - \mathbf{g}_t \|^2 - \frac{\eta}{2} \left( \|\nabla L_i(\theta_{t-1})\|^2 + (1 - \eta L) \|\mathbf{g}_t\|^2 \right)\nonumber \\
&\overset{(c)}{\leq} \frac{\eta}{2} \left( (1 - b_t) \|\nabla L_i(\theta_{t-1}) - \mathbf{g}_t^a \|^2 + b_t \|\nabla L_i(\theta_{t-1}) - \mathbf{g}_t^b \|^2 \right) - \frac{\eta}{2} \left( \|\nabla L_i(\theta_{t-1})\|^2 + (1 - \eta L) \|\mathbf{g}_t\|^2 \right) \nonumber\\
&\overset{(d)}{\leq} \frac{\eta}{2} \left( (1 - b_t) \|\nabla L_i(\theta_{t-1}) - \mathbf{g}_t^a \|^2 + 4b_t G^2 \right) - \eta \mu L_i(\theta_{t-1}),
\end{align}
where $(a)$ is due to Assumption 2-(i); 
$(b)$ follows the update of $\theta_{t} = \theta_{t-1} - \eta\mathbf{g}_{t}$; 
$(c)$ is due to the definition of $\mathbf{g}_t$ and the convexity of $\| \cdot \|^2$; 
$(d)$ follows the Assumption 2-(i),(ii), and $\eta L \le 1$. 

Let $\mathcal{F}_t^{\tau}:=\mathcal{F}\left(I(X_j\in\mathcal{A}_t^{\tau}),j\in[n_t]\right)$. 
Then we have 
\begin{align*}
\E_* \left( \left\| \mathbf{g}_t^a - \nabla L_i(\theta_{t-1}) \right\|^2 \big| \mathcal{F}_t^{\tau}\right) &= 
\frac{1}{|\mathcal{A}_t^{\tau}|^2} \E_* \left(\Bigl\|  \sum_{X \in \mathcal{A}_t^{\tau}} \left( \nabla l(\theta_t; X) - \nabla L(\theta_t) \right) \Bigr\|^2 \bigg| \mathcal{F}_t^{\tau}\right) \\
&= \frac{1}{|\mathcal{A}_t^{\tau}|^2}  \E_* \left(\sum_{X \in \mathcal{A}_t^{\tau}}\Bigl\|  \nabla l(\theta_t; X) - \nabla L(\theta_t) \Bigr\|^2 \bigg| \mathcal{F}_t^{\tau}\right) \\
&\leq \frac{4G^2}{|\mathcal{A}_t^{\tau}|}, 
\end{align*}

where the last inequality holds due to Assumption 2-(i).
For a given $\delta>0$, by using Lemma \ref{lem:3} with $\lambda=1/\delta$, we have 
\begin{align*}
P_*\left( P_* \biggl(\left\| \mathbf{g}_t^a - \nabla L_i(\theta_{t-1}) \right\|^2  \le \frac{4G^2}{\delta|\mathcal{A}_t^{\tau}|}  \bigg| \mathcal{F}_t^{\tau} \biggr)\ge 1-\delta \right)=1,
\end{align*}
and hence
\begin{align}
\label{ineq10}
 P_* \biggl(\left\| \mathbf{g}_t^a - \nabla L_i(\theta_{t-1}) \right\|^2  \le \frac{4G^2}{\delta|\mathcal{A}_t^{\tau}|}   \biggr)\ge 1-\delta. 
\end{align} 
We combine (\ref{ineq9}) and (\ref{ineq10}) to achieve 
\begin{align}
\label{ineq11}
\left\|\mathbf{g}_t^a - \nabla L_i(\theta_{t-1}) \right\|^2 \le  \frac{4G^2}{\delta c_1 n_0}\gamma^{-(t-1)},
\end{align} 
with a probability of at least $1-5\delta$. 

Using the above bound in (\ref{ineq11}), we can further expand the bound in (\ref{ineq12}) as follows:
\begin{align*}
L_i(\theta_{t}) - L_i(\theta_{t-1}) &\leq \frac{\eta}{2} \left( (1 - b_t) \frac{4G^2}{\delta c_1 n_0}\gamma^{-(t-1)} + 4b_t G^2 \right) - \eta \mu L_i(\theta_{t-1}) \\
&\leq \frac{\eta}{2} \left( \frac{4G^2}{\delta c_1 n_0}\gamma^{-(t-1)} + 4G^2\frac{c_2}{c_1} \gamma^{-(t-1)} \right) - \eta \mu L_i(\theta_{t-1}) \\
&= \frac{2 \eta G^2}{c_1} \left( \frac{1}{\delta n_0} + c_2 \right) \gamma^{-(t-1)} - \eta \mu L_i(\theta_{t-1}).
\end{align*}

Hence, we have 
\begin{align*}
    L_i(\theta_{t}) &\le (1-\eta\mu)L_i(\theta_{t-1})+\frac{2 \eta G^2}{c_1} \left( \frac{1}{\delta n_0} + c_2 \right)\gamma^{-(t-1)} \\ 
    &\le \gamma\left[ (1-\eta\mu)a_2 + \frac{2 \eta G^2}{c_1} \left( \frac{1}{\delta n_0} + c_2 \right)\right] \gamma^{-t},
\end{align*} 
with a probability of at least $1-5\delta$.
Let us select the learning rate $\eta$ between the interval given as:
\begin{align*}
\frac{1}{\mu}\left( 1-\frac{1}{2\gamma}  \right) \le \eta \le \frac{a_2 c_1}{4\gamma G^2\left(\frac{1}{\delta n_0}+c_2\right)}.
\end{align*}
Then, we have 
\begin{align*}
    L_i(\theta_{t}) &\le \gamma\left[ (1-\eta\mu)a_2 + \frac{2 \eta G^2}{c_1} \left( \frac{1}{\delta n_0} + c_2 \right)\right] \gamma^{-t}\\
    &\le \gamma \left( \frac{a_2}{2\gamma} + \frac{a_2}{2\gamma} \right) \gamma^{-t}\\
    &=a_2 \gamma^{-t},
\end{align*} 
and the proof is completed. \qed

\section{Detailed experiment results}

\subsection{Description of Imporance weighted autoencoders (IWAE)}

Importance Weighted Autoencoders (IWAE) \citep{DBLP:journals/corr/BurdaGS15} is a variational inference method designed to produce arbitrarily tight lower bounds. For a given input \(\boldsymbol{x}\), IWAE minimizes the following expression, which leverages multiple samples from the variational distribution \(q(\boldsymbol{z}|\boldsymbol{x};\phi)\):

\begin{align*}
-\mathbb{E}_{\boldsymbol{z}_1,...,\boldsymbol{z}_K\sim q(\boldsymbol{z}|\boldsymbol{x};\phi)}\left[ \log \left( \frac{1}{K} \sum_{k=1}^K \frac{p(\boldsymbol{x},\boldsymbol{z}_k;\theta)}{q(\boldsymbol{z}_k|\boldsymbol{x};\phi)} \right) \right],
\end{align*}
where \(K\) is the number of samples. In practice, IWAE approximates this lower bound using the Monte Carlo method, which is expressed as:
\begin{align*}
\hat{L}^{\text{IWAE}}(\theta,\phi;\boldsymbol{x}) := - \log \left( \frac{1}{K} \sum_{k=1}^K \frac{p(\boldsymbol{x},\boldsymbol{z}_k;\theta)}{q(\boldsymbol{z}_k|\boldsymbol{x};\phi)} \right),
\end{align*}
where \(\boldsymbol{z}_k \sim q(\boldsymbol{z}|\boldsymbol{x};\phi)\) for \(k=1,\dots,K\).
As for the encoder and decoder, as done in \citet{kim2024odimoutlierdetectionlikelihood}, we use two hidden layered DNN architectures with 50 to 100 hidden nodes for each hidden layer. 
And we set $K$, the number of samples drawn from the encoder used for constructing the IWAE objective function, to two. 
We minimize the above loss function with respect to $(\theta,\phi)$ using an SGD-based optimizer such as Adam \citep{kingma2014adam}.

\clearpage

\subsection{Data description }
We evaluate a total of 46 tabular datasets, 6 image datasets, and 5 text datasets. 
These datasets are all obtained from a source known as ADBench. (Han et al., 2022b). Table \ref{tab:adbench_summ} provides a summary of the basic information for all datasets we analyze.
\begin{table}[h!]
\renewcommand\thetable{B.1}
\centering
\setlength{\tabcolsep}{1mm}
\fontsize{8pt}{8pt}\selectfont
\begin{tabular}{llccccl}
\toprule
\textbf{Number} & \textbf{Dataset Name} & \textbf{\#Samples} & \textbf{\#Features} & \textbf{\#Anomaly} & \textbf{\%Anomaly} & \textbf{Category} \\
\midrule
\textbf{1}      & ALOI                  & 49534              & 27                  & 1508               & 3.04               & Image             \\
\textbf{2}      & annthyroid            & 7200               & 6                   & 534                & 7.42               & Healthcare        \\
\textbf{3}      & backdoor              & 95329              & 196                 & 2329               & 2.44               & Network           \\
\textbf{4}      & breastw               & 683                & 9                   & 239                & 34.99              & Healthcare        \\
\textbf{5}      & campaign              & 41188              & 62                  & 4640               & 11.27              & Finance           \\
\textbf{6}      & cardio                & 1831               & 21                  & 176                & 9.61               & Healthcare        \\
\textbf{7}      & Cardiotocography      & 2114               & 21                  & 466                & 22.04              & Healthcare        \\
\textbf{8}      & celeba                & 202599             & 39                  & 4547               & 2.24               & Image             \\
\textbf{9}      & census                & 299285             & 500                 & 18568              & 6.2                & Sociology         \\
\textbf{10}     & cover                 & 286048             & 10                  & 2747               & 0.96               & Botany            \\
\textbf{11}     & donors                & 619326             & 10                  & 36710              & 5.93               & Sociology         \\
\textbf{12}     & fault                 & 1941               & 27                  & 673                & 34.67              & Physical          \\
\textbf{13}     & fraud                 & 284807             & 29                  & 492                & 0.17               & Finance           \\
\textbf{14}     & glass                 & 214                & 7                   & 9                  & 4.21               & Forensic          \\
\textbf{15}     & Hepatitis             & 80                 & 19                  & 13                 & 16.25              & Healthcare        \\
\textbf{16}     & http                  & 567498             & 3                   & 2211               & 0.39               & Web               \\
\textbf{17}     & InternetAds           & 1966               & 1555                & 368                & 18.72              & Image             \\
\textbf{18}     & Ionosphere            & 351                & 32                  & 126                & 35.9               & Oryctognosy       \\
\textbf{19}     & landsat               & 6435               & 36                  & 1333               & 20.71              & Astronautics      \\
\textbf{20}     & letter                & 1600               & 32                  & 100                & 6.25               & Image             \\
\textbf{21}     & Lymphography          & 148                & 18                  & 6                  & 4.05               & Healthcare        \\
\textbf{22}     & magic.gamma           & 19020              & 10                  & 6688               & 35.16              & Physical          \\
\textbf{23}     & mammography           & 11183              & 6                   & 260                & 2.32               & Healthcare        \\
\textbf{24}     & mnist                 & 7603               & 100                 & 700                & 9.21               & Image             \\
\textbf{25}     & musk                  & 3062               & 166                 & 97                 & 3.17               & Chemistry         \\
\textbf{26}     & optdigits             & 5216               & 64                  & 150                & 2.88               & Image             \\
\textbf{27}     & PageBlocks            & 5393               & 10                  & 510                & 9.46               & Document          \\
\textbf{28}     & pendigits             & 6870               & 16                  & 156                & 2.27               & Image             \\
\textbf{29}     & Pima                  & 768                & 8                   & 268                & 34.9               & Healthcare        \\
\textbf{30}     & satellite             & 6435               & 36                  & 2036               & 31.64              & Astronautics      \\
\textbf{31}     & satimage-2            & 5803               & 36                  & 71                 & 1.22               & Astronautics      \\
\textbf{32}     & shuttle               & 49097              & 9                   & 3511               & 7.15               & Astronautics      \\
\textbf{33}     & skin                  & 245057             & 3                   & 50859              & 20.75              & Image             \\
\textbf{34}     & smtp                  & 95156              & 3                   & 30                 & 0.03               & Web               \\
\textbf{35}     & SpamBase              & 4207               & 57                  & 1679               & 39.91              & Document          \\
\textbf{36}     & speech                & 3686               & 400                 & 61                 & 1.65               & Linguistics       \\
\textbf{37}     & Stamps                & 340                & 9                   & 31                 & 9.12               & Document          \\
\textbf{38}     & thyroid               & 3772               & 6                   & 93                 & 2.47               & Healthcare        \\
\textbf{39}     & vertebral             & 240                & 6                   & 30                 & 12.5               & Biology           \\
\textbf{40}     & vowels                & 1456               & 12                  & 50                 & 3.43               & Linguistics       \\
\textbf{41}     & Waveform              & 3443               & 21                  & 100                & 2.9                & Physics           \\
\textbf{42}     & WBC                   & 223                & 9                   & 10                 & 4.48               & Healthcare        \\
\textbf{43}     & WDBC                  & 367                & 30                  & 10                 & 2.72               & Healthcare        \\
\textbf{44}     & Wilt                  & 4819               & 5                   & 257                & 5.33               & Botany            \\
\textbf{45}     & wine                  & 129                & 13                  & 10                 & 7.75               & Chemistry         \\
\textbf{46}     & WPBC                  & 198                & 33                  & 47                 & 23.74              & Healthcare        \\
\textbf{47}     & yeast                 & 1484               & 8                   & 507                & 34.16              & Biology           \\
\textbf{48}     & CIFAR10               & 5263               & 512                 & 263                & 5                  & Image             \\
\textbf{49}     & FashionMNIST          & 6315               & 512                 & 315                & 5                  & Image             \\
\textbf{50}     & MNIST-C               & 10000              & 512                 & 500                & 5                  & Image             \\
\textbf{51}     & MVTec-AD              & 5354               & 512                 & 1258               & 23.5               & Image             \\
\textbf{52}     & SVHN                  & 5208               & 512                 & 260                & 5                  & Image             \\
\textbf{53}     & Agnews                & 10000              & 768                 & 500                & 5                  & NLP               \\
\textbf{54}     & Amazon                & 10000              & 768                 & 500                & 5                  & NLP               \\
\textbf{55}     & Imdb                  & 10000              & 768                 & 500                & 5                  & NLP               \\
\textbf{56}     & Yelp                  & 10000              & 768                 & 500                & 5                  & NLP               \\
\textbf{57}     & 20news        & 11905              & 768                 & 591                & 4.96               & NLP              \\
\bottomrule
\end{tabular}
\caption{Description of \texttt{ADBench} datasets}
\label{tab:adbench_summ}
\end{table}

\subsection{Detailed AUC and PRAUC results over ADBench datasets }
Table \ref{tab:adbench1}- \ref{tab:adbench4} provide detailed results of averaged AUC and PRAUC for each method over the \texttt{ADBench} datasets in unsupervised and semi-supervised settings.
\begin{sidewaystable}[h!]
\renewcommand\thetable{B.2}
\centering
\setlength{\tabcolsep}{0.7mm} 
\fontsize{6pt}{7pt}\selectfont 
\begin{tabular}{lccccccccccccccccccccccccc}
\toprule
    & CBLOF & COPOD & ECOD  & FeatureBagging & HBOS  & IForest & kNN   & LODA  & LOF   & MCD   & OCSVM & PCA   & DAGMM & DeepSVDD & DROCC & GOAD  & ICL   & PlanarFlow & DDPM  & DTE-NP & DTE-IG & DTE-C & ODIM  & ALTBI-G & ALTBI-I \\
\midrule
aloi             & 0.556 & 0.515 & 0.531 & 0.792          & 0.531 & 0.542   & 0.613 & 0.495 & 0.767 & 0.520 & 0.549 & 0.549 & 0.517 & 0.514    & 0.500 & 0.497 & 0.548 & 0.520      & 0.532 & 0.645  & 0.541  & 0.525 & 0.527 & 0.545    & 0.534    \\
annthyroid       & 0.676 & 0.777 & 0.789 & 0.788          & 0.608 & 0.816   & 0.761 & 0.453 & 0.710 & 0.918 & 0.682 & 0.676 & 0.548 & 0.739    & 0.631 & 0.453 & 0.599 & 0.966      & 0.814 & 0.781  & 0.923  & 0.964 & 0.603 & 0.864    & 0.642    \\
backdoor         & 0.897 & 0.500 & 0.500 & 0.790          & 0.740 & 0.725   & 0.826 & 0.515 & 0.764 & 0.848 & 0.889 & 0.888 & 0.752 & 0.735    & 0.500 & 0.587 & 0.936 & 0.787      & 0.892 & 0.806  & 0.753  & 0.875 & 0.886 & 0.881    & 0.873    \\
breastw          & 0.961 & 0.994 & 0.990 & 0.408          & 0.984 & 0.983   & 0.980 & 0.970 & 0.446 & 0.985 & 0.935 & 0.946 & 0.811 & 0.625    & 0.847 & 0.845 & 0.807 & 0.965      & 0.766 & 0.976  & 0.905  & 0.891 & 0.992 & 0.989    & 0.981    \\
campaign         & 0.738 & 0.783 & 0.769 & 0.594          & 0.768 & 0.704   & 0.750 & 0.493 & 0.614 & 0.775 & 0.737 & 0.734 & 0.581 & 0.508    & 0.500 & 0.443 & 0.766 & 0.566      & 0.724 & 0.746  & 0.660  & 0.789 & 0.727 & 0.746    & 0.725    \\
cardio           & 0.832 & 0.921 & 0.935 & 0.579          & 0.839 & 0.922   & 0.830 & 0.856 & 0.551 & 0.815 & 0.934 & 0.949 & 0.625 & 0.498    & 0.655 & 0.908 & 0.461 & 0.796      & 0.723 & 0.777  & 0.631  & 0.721 & 0.911 & 0.714    & 0.857    \\
cardiotocography & 0.561 & 0.664 & 0.784 & 0.538          & 0.595 & 0.681   & 0.503 & 0.708 & 0.527 & 0.500 & 0.691 & 0.747 & 0.546 & 0.488    & 0.449 & 0.624 & 0.372 & 0.643      & 0.579 & 0.493  & 0.506  & 0.510 & 0.649 & 0.483    & 0.542    \\
celeba           & 0.753 & 0.757 & 0.763 & 0.514          & 0.754 & 0.707   & 0.736 & 0.600 & 0.432 & 0.803 & 0.781 & 0.792 & 0.627 & 0.491    & 0.726 & 0.432 & 0.684 & 0.703      & 0.796 & 0.699  & 0.700  & 0.812 & 0.839 & 0.724    & 0.803    \\
census           & 0.664 & 0.500 & 0.500 & 0.538          & 0.611 & 0.607   & 0.671 & 0.454 & 0.562 & 0.731 & 0.655 & 0.662 & 0.491 & 0.527    & 0.443 & 0.488 & 0.668 & 0.604      & 0.659 & 0.672  & 0.629  & 0.646 & 0.665 & 0.702    & 0.662    \\
cover            & 0.922 & 0.882 & 0.919 & 0.571          & 0.707 & 0.873   & 0.866 & 0.922 & 0.568 & 0.696 & 0.952 & 0.934 & 0.742 & 0.580    & 0.747 & 0.124 & 0.681 & 0.417      & 0.808 & 0.838  & 0.635  & 0.697 & 0.901 & 0.924    & 0.901    \\
donors           & 0.808 & 0.815 & 0.888 & 0.691          & 0.743 & 0.771   & 0.829 & 0.566 & 0.629 & 0.765 & 0.770 & 0.825 & 0.558 & 0.511    & 0.747 & 0.225 & 0.739 & 0.899      & 0.806 & 0.832  & 0.796  & 0.785 & 0.813 & 0.810    & 0.601    \\
fault            & 0.665 & 0.455 & 0.468 & 0.591          & 0.506 & 0.544   & 0.715 & 0.478 & 0.579 & 0.505 & 0.537 & 0.480 & 0.495 & 0.522    & 0.668 & 0.546 & 0.661 & 0.469      & 0.562 & 0.726  & 0.577  & 0.590 & 0.544 & 0.703    & 0.664    \\
fraud            & 0.954 & 0.943 & 0.949 & 0.616          & 0.945 & 0.950   & 0.955 & 0.856 & 0.548 & 0.911 & 0.954 & 0.952 & 0.857 & 0.769    & 0.500 & 0.724 & 0.931 & 0.895      & 0.924 & 0.956  & 0.942  & 0.938 & 0.940 & 0.957    & 0.923    \\
glass            & 0.855 & 0.760 & 0.710 & 0.659          & 0.820 & 0.790   & 0.870 & 0.624 & 0.618 & 0.795 & 0.661 & 0.715 & 0.630 & 0.517    & 0.743 & 0.545 & 0.729 & 0.766      & 0.560 & 0.881  & 0.681  & 0.864 & 0.708 & 0.812    & 0.771    \\
hepatitis        & 0.635 & 0.807 & 0.737 & 0.469          & 0.768 & 0.683   & 0.669 & 0.557 & 0.468 & 0.721 & 0.704 & 0.748 & 0.600 & 0.361    & 0.582 & 0.637 & 0.616 & 0.654      & 0.461 & 0.631  & 0.451  & 0.577 & 0.781 & 0.674    & 0.804    \\
http             & 0.996 & 0.991 & 0.980 & 0.288          & 0.991 & 1.000   & 0.051 & 0.060 & 0.338 & 1.000 & 0.994 & 0.997 & 0.838 & 0.249    & 0.500 & 0.996 & 0.921 & 0.994      & 0.998 & 0.051  & 0.973  & 0.995 & 0.995 & 0.994    & 0.996    \\
internetads      & 0.616 & 0.676 & 0.677 & 0.494          & 0.696 & 0.686   & 0.616 & 0.541 & 0.587 & 0.660 & 0.615 & 0.609 & 0.515 & 0.583    & 0.500 & 0.614 & 0.592 & 0.608      & 0.614 & 0.634  & 0.635  & 0.656 & 0.618 & 0.690    & 0.720    \\
ionosphere       & 0.892 & 0.783 & 0.717 & 0.876          & 0.544 & 0.833   & 0.922 & 0.788 & 0.864 & 0.951 & 0.838 & 0.777 & 0.641 & 0.514    & 0.766 & 0.829 & 0.629 & 0.884      & 0.758 & 0.924  & 0.697  & 0.911 & 0.768 & 0.922    & 0.857    \\
landsat          & 0.548 & 0.422 & 0.368 & 0.540          & 0.575 & 0.474   & 0.614 & 0.382 & 0.549 & 0.607 & 0.423 & 0.366 & 0.533 & 0.631    & 0.626 & 0.506 & 0.649 & 0.464      & 0.496 & 0.602  & 0.473  & 0.544 & 0.457 & 0.571    & 0.538    \\
letter           & 0.763 & 0.560 & 0.573 & 0.886          & 0.589 & 0.616   & 0.812 & 0.537 & 0.878 & 0.804 & 0.598 & 0.524 & 0.503 & 0.517    & 0.780 & 0.598 & 0.737 & 0.689      & 0.847 & 0.850  & 0.676  & 0.781 & 0.628 & 0.906    & 0.744    \\
lymphography     & 0.994 & 0.996 & 0.995 & 0.523          & 0.995 & 0.999   & 0.995 & 0.900 & 0.636 & 0.989 & 0.996 & 0.997 & 0.840 & 0.681    & 0.878 & 0.995 & 0.884 & 0.940      & 0.958 & 0.989  & 0.852  & 0.834 & 1.000 & 0.987    & 0.991    \\
magic.gamma      & 0.725 & 0.681 & 0.638 & 0.700          & 0.709 & 0.721   & 0.795 & 0.655 & 0.678 & 0.699 & 0.673 & 0.667 & 0.584 & 0.604    & 0.728 & 0.442 & 0.676 & 0.742      & 0.763 & 0.801  & 0.782  & 0.765 & 0.728 & 0.829    & 0.743    \\
mammography      & 0.795 & 0.905 & 0.906 & 0.726          & 0.838 & 0.860   & 0.852 & 0.867 & 0.702 & 0.690 & 0.871 & 0.888 & 0.719 & 0.451    & 0.779 & 0.414 & 0.658 & 0.782      & 0.749 & 0.849  & 0.799  & 0.810 & 0.835 & 0.865    & 0.806    \\
musk             & 1.000 & 0.948 & 0.953 & 0.575          & 1.000 & 0.998   & 0.964 & 0.993 & 0.581 & 1.000 & 1.000 & 1.000 & 0.912 & 0.538    & 0.575 & 1.000 & 0.790 & 0.748      & 1.000 & 0.882  & 0.785  & 0.965 & 1.000 & 1.000    & 1.000    \\
optdigits        & 0.785 & 0.500 & 0.500 & 0.539          & 0.868 & 0.696   & 0.395 & 0.493 & 0.538 & 0.413 & 0.507 & 0.518 & 0.408 & 0.519    & 0.565 & 0.657 & 0.533 & 0.492      & 0.402 & 0.386  & 0.513  & 0.508 & 0.570 & 0.536    & 0.684    \\
pageblocks       & 0.893 & 0.875 & 0.914 & 0.758          & 0.779 & 0.897   & 0.919 & 0.712 & 0.703 & 0.923 & 0.914 & 0.907 & 0.753 & 0.592    & 0.914 & 0.609 & 0.768 & 0.908      & 0.820 & 0.906  & 0.850  & 0.924 & 0.886 & 0.887    & 0.868    \\
pendigits        & 0.864 & 0.906 & 0.927 & 0.518          & 0.925 & 0.947   & 0.828 & 0.895 & 0.534 & 0.834 & 0.929 & 0.936 & 0.548 & 0.383    & 0.520 & 0.592 & 0.650 & 0.780      & 0.700 & 0.786  & 0.624  & 0.713 & 0.945 & 0.892    & 0.954    \\
pima             & 0.655 & 0.662 & 0.604 & 0.573          & 0.704 & 0.674   & 0.723 & 0.595 & 0.563 & 0.686 & 0.631 & 0.651 & 0.522 & 0.510    & 0.542 & 0.606 & 0.524 & 0.615      & 0.537 & 0.707  & 0.599  & 0.624 & 0.689 & 0.713    & 0.717    \\
satellite        & 0.742 & 0.633 & 0.583 & 0.545          & 0.762 & 0.695   & 0.721 & 0.614 & 0.550 & 0.804 & 0.662 & 0.601 & 0.675 & 0.562    & 0.608 & 0.702 & 0.627 & 0.671      & 0.715 & 0.702  & 0.582  & 0.711 & 0.712 & 0.677    & 0.760    \\
satimage-2       & 0.999 & 0.975 & 0.965 & 0.526          & 0.976 & 0.993   & 0.992 & 0.981 & 0.539 & 0.995 & 0.997 & 0.977 & 0.911 & 0.551    & 0.579 & 0.996 & 0.898 & 0.970      & 0.996 & 0.980  & 0.858  & 0.946 & 0.998 & 0.998    & 0.998    \\
shuttle          & 0.621 & 0.995 & 0.993 & 0.493          & 0.986 & 0.997   & 0.732 & 0.389 & 0.526 & 0.990 & 0.992 & 0.990 & 0.898 & 0.576    & 0.500 & 0.208 & 0.642 & 0.852      & 0.975 & 0.698  & 0.669  & 0.976 & 0.985 & 0.996    & 0.982    \\
skin             & 0.675 & 0.471 & 0.490 & 0.534          & 0.588 & 0.670   & 0.720 & 0.442 & 0.550 & 0.892 & 0.547 & 0.447 & 0.554 & 0.548    & 0.708 & 0.579 & 0.265 & 0.773      & 0.461 & 0.718  & 0.741  & 0.741 & 0.626 & 0.812    & 0.874    \\
smtp             & 0.863 & 0.912 & 0.882 & 0.794          & 0.809 & 0.905   & 0.933 & 0.819 & 0.899 & 0.948 & 0.845 & 0.856 & 0.868 & 0.895    & 0.500 & 0.915 & 0.656 & 0.784      & 0.956 & 0.930  & 0.769  & 0.951 & 0.844 & 0.872    & 0.846    \\
spambase         & 0.541 & 0.688 & 0.656 & 0.424          & 0.664 & 0.637   & 0.566 & 0.480 & 0.453 & 0.446 & 0.534 & 0.548 & 0.488 & 0.584    & 0.490 & 0.496 & 0.459 & 0.528      & 0.510 & 0.545  & 0.509  & 0.515 & 0.555 & 0.655    & 0.503    \\
speech           & 0.471 & 0.489 & 0.470 & 0.509          & 0.473 & 0.476   & 0.480 & 0.466 & 0.512 & 0.494 & 0.466 & 0.469 & 0.522 & 0.512    & 0.483 & 0.458 & 0.512 & 0.496      & 0.466 & 0.487  & 0.488  & 0.495 & 0.465 & 0.457    & 0.477    \\
stamps           & 0.660 & 0.929 & 0.877 & 0.502          & 0.904 & 0.907   & 0.870 & 0.831 & 0.512 & 0.838 & 0.882 & 0.909 & 0.719 & 0.465    & 0.760 & 0.774 & 0.505 & 0.838      & 0.556 & 0.820  & 0.692  & 0.753 & 0.920 & 0.646    & 0.883    \\
thyroid          & 0.909 & 0.939 & 0.977 & 0.707          & 0.948 & 0.979   & 0.965 & 0.819 & 0.657 & 0.986 & 0.958 & 0.955 & 0.719 & 0.505    & 0.889 & 0.574 & 0.693 & 0.992      & 0.871 & 0.964  & 0.828  & 0.990 & 0.917 & 0.977    & 0.947    \\
vertebral        & 0.463 & 0.263 & 0.417 & 0.473          & 0.317 & 0.362   & 0.379 & 0.294 & 0.487 & 0.389 & 0.426 & 0.378 & 0.470 & 0.394    & 0.425 & 0.468 & 0.449 & 0.409      & 0.563 & 0.400  & 0.451  & 0.458 & 0.316 & 0.327    & 0.402    \\
vowels           & 0.884 & 0.496 & 0.593 & 0.933          & 0.679 & 0.763   & 0.951 & 0.705 & 0.932 & 0.732 & 0.779 & 0.604 & 0.464 & 0.514    & 0.738 & 0.791 & 0.784 & 0.888      & 0.903 & 0.964  & 0.705  & 0.914 & 0.843 & 0.955    & 0.868    \\
waveform         & 0.701 & 0.739 & 0.603 & 0.715          & 0.694 & 0.707   & 0.750 & 0.594 & 0.693 & 0.572 & 0.669 & 0.635 & 0.523 & 0.609    & 0.674 & 0.592 & 0.661 & 0.640      & 0.617 & 0.729  & 0.523  & 0.602 & 0.700 & 0.678    & 0.746    \\
wbc              & 0.977 & 0.994 & 0.994 & 0.388          & 0.987 & 0.996   & 0.982 & 0.992 & 0.607 & 0.988 & 0.987 & 0.993 & 0.821 & 0.503    & 0.821 & 0.949 & 0.853 & 0.934      & 0.948 & 0.979  & 0.894  & 0.871 & 1.000 & 0.983    & 0.995    \\
wdbc             & 0.990 & 0.993 & 0.971 & 0.867          & 0.989 & 0.988   & 0.980 & 0.980 & 0.849 & 0.969 & 0.984 & 0.988 & 0.715 & 0.602    & 0.347 & 0.983 & 0.738 & 0.985      & 0.965 & 0.975  & 0.566  & 0.835 & 0.965 & 0.962    & 0.979    \\
wilt             & 0.396 & 0.345 & 0.394 & 0.666          & 0.348 & 0.451   & 0.511 & 0.313 & 0.678 & 0.859 & 0.317 & 0.239 & 0.432 & 0.465    & 0.400 & 0.555 & 0.649 & 0.794      & 0.659 & 0.552  & 0.834  & 0.851 & 0.322 & 0.644    & 0.357    \\
wine             & 0.453 & 0.865 & 0.738 & 0.323          & 0.907 & 0.786   & 0.470 & 0.822 & 0.330 & 0.975 & 0.671 & 0.819 & 0.513 & 0.507    & 0.621 & 0.734 & 0.455 & 0.390      & 0.374 & 0.425  & 0.310  & 0.557 & 0.904 & 0.825    & 0.932    \\
wpbc             & 0.487 & 0.519 & 0.489 & 0.436          & 0.548 & 0.516   & 0.512 & 0.501 & 0.447 & 0.534 & 0.485 & 0.486 & 0.449 & 0.493    & 0.483 & 0.466 & 0.488 & 0.483      & 0.493 & 0.502  & 0.489  & 0.468 & 0.544 & 0.539    & 0.495    \\
yeast            & 0.461 & 0.380 & 0.443 & 0.465          & 0.402 & 0.394   & 0.396 & 0.461 & 0.453 & 0.406 & 0.420 & 0.418 & 0.503 & 0.520    & 0.396 & 0.503 & 0.466 & 0.442      & 0.463 & 0.400  & 0.446  & 0.420 & 0.398 & 0.419    & 0.411    \\
CIFAR10          & 0.663 & 0.548 & 0.567 & 0.687          & 0.572 & 0.629   & 0.659 & 0.591 & 0.686 & 0.639 & 0.663 & 0.659 & 0.530 & 0.555    & 0.503 & 0.659 & 0.557 & 0.621      & 0.663 & 0.660  & 0.595  & 0.629 & 0.928 & 0.913    & 0.888    \\
MNIST-C          & 0.757 & 0.500 & 0.500 & 0.702          & 0.689 & 0.733   & 0.786 & 0.591 & 0.699 & 0.739 & 0.751 & 0.741 & 0.581 & 0.552    & 0.594 & 0.752 & 0.670 & 0.705      & 0.751 & 0.788  & 0.703  & 0.746 & 0.737 & 0.743    & 0.740    \\
MVTec-AD         & 0.754 & 0.500 & 0.500 & 0.745          & 0.732 & 0.747   & 0.763 & 0.644 & 0.742 & 0.618 & 0.735 & 0.724 & 0.596 & 0.603    & 0.544 & 0.730 & 0.683 & 0.637      & 0.732 & 0.761  & 0.655  & 0.730 & 0.941 & 0.850    & 0.880    \\
SVHN             & 0.601 & 0.500 & 0.500 & 0.629          & 0.542 & 0.580   & 0.604 & 0.534 & 0.628 & 0.583 & 0.604 & 0.599 & 0.528 & 0.521    & 0.521 & 0.597 & 0.571 & 0.580      & 0.605 & 0.607  & 0.567  & 0.600 & 0.553 & 0.541    & 0.581    \\
mnist            & 0.843 & 0.500 & 0.500 & 0.664          & 0.574 & 0.811   & 0.867 & 0.564 & 0.658 & 0.856 & 0.849 & 0.848 & 0.631 & 0.605    & 0.615 & 0.698 & 0.691 & 0.645      & 0.816 & 0.853  & 0.756  & 0.819 & 0.859 & 0.545    & 0.866    \\
FashionMNIST     & 0.871 & 0.500 & 0.500 & 0.748          & 0.748 & 0.831   & 0.875 & 0.672 & 0.738 & 0.840 & 0.860 & 0.853 & 0.664 & 0.647    & 0.564 & 0.860 & 0.758 & 0.819      & 0.861 & 0.873  & 0.767  & 0.841 & 0.907 & 0.631    & 0.879    \\
20news           & 0.564 & 0.533 & 0.544 & 0.610          & 0.537 & 0.550   & 0.567 & 0.539 & 0.610 & 0.583 & 0.559 & 0.545 & 0.518 & 0.515    & 0.496 & 0.553 & 0.547 & 0.513      & 0.547 & 0.570  & 0.527  & 0.579 & 0.716 & 0.675    & 0.722    \\
agnews           & 0.619 & 0.551 & 0.552 & 0.715          & 0.554 & 0.584   & 0.647 & 0.568 & 0.714 & 0.665 & 0.601 & 0.566 & 0.508 & 0.494    & 0.500 & 0.592 & 0.591 & 0.497      & 0.571 & 0.652  & 0.545  & 0.627 & 0.788 & 0.728    & 0.868    \\
amazon           & 0.579 & 0.571 & 0.541 & 0.572          & 0.563 & 0.558   & 0.603 & 0.526 & 0.571 & 0.597 & 0.565 & 0.550 & 0.501 & 0.464    & 0.500 & 0.560 & 0.528 & 0.495      & 0.551 & 0.603  & 0.535  & 0.556 & 0.520 & 0.520    & 0.546    \\
imdb             & 0.496 & 0.512 & 0.471 & 0.499          & 0.499 & 0.489   & 0.494 & 0.466 & 0.500 & 0.504 & 0.484 & 0.478 & 0.487 & 0.526    & 0.500 & 0.486 & 0.521 & 0.493      & 0.478 & 0.495  & 0.486  & 0.484 & 0.508 & 0.488    & 0.528    \\
yelp             & 0.635 & 0.605 & 0.578 & 0.661          & 0.600 & 0.602   & 0.670 & 0.581 & 0.661 & 0.655 & 0.621 & 0.592 & 0.498 & 0.524    & 0.504 & 0.590 & 0.545 & 0.527      & 0.594 & 0.671  & 0.514  & 0.602 & 0.551 & 0.556    & 0.556    \\
\midrule
\textbf{Average} & 0.731 & 0.692 & 0.689 & 0.608          & 0.714 & 0.737   & 0.728 & 0.627 & 0.611 & 0.748 & 0.725 & 0.718 & 0.614 & 0.543    & 0.600 & 0.627 & 0.643 & 0.696      & 0.700 & 0.722  & 0.660  & 0.728 & 0.748 & 0.753    & 0.762     \\ 
\bottomrule
\end{tabular}
\caption{ROC AUC for the unsupervised setting on ADBench}
\label{tab:adbench1}
\end{sidewaystable}

\begin{sidewaystable}[h!]
\renewcommand\thetable{B.3}
\centering
\setlength{\tabcolsep}{0.7mm} 
\fontsize{6pt}{7pt}\selectfont 
\begin{tabular}{lccccccccccccccccccccccccc}
\toprule
    & CBLOF & COPOD & ECOD  & FeatureBagging & HBOS  & IForest & kNN   & LODA  & LOF   & MCD   & OCSVM & PCA   & DAGMM & DeepSVDD & DROCC & GOAD  & ICL   & PlanarFlow & DDPM  & DTE-NP & DTE-IG & DTE-C & ODIM  & ALTBI-G & ALTBI-I  \\ 
\midrule
aloi             & 0.037 & 0.031 & 0.033 & 0.104          & 0.034 & 0.034   & 0.048 & 0.033 & 0.097 & 0.032 & 0.039 & 0.037 & 0.033 & 0.034    & 0.030 & 0.033 & 0.046 & 0.032      & 0.036 & 0.056  & 0.040  & 0.033 & 0.040 & 0.037   & 0.036   \\
annthyroid       & 0.169 & 0.174 & 0.272 & 0.206          & 0.228 & 0.312   & 0.224 & 0.098 & 0.163 & 0.503 & 0.188 & 0.196 & 0.109 & 0.192    & 0.186 & 0.131 & 0.123 & 0.654      & 0.297 & 0.228  & 0.380  & 0.670 & 0.166 & 0.350   & 0.156   \\
backdoor         & 0.547 & 0.025 & 0.025 & 0.217          & 0.052 & 0.045   & 0.479 & 0.101 & 0.358 & 0.122 & 0.534 & 0.531 & 0.250 & 0.372    & 0.025 & 0.347 & 0.717 & 0.336      & 0.520 & 0.473  & 0.438  & 0.481 & 0.406 & 0.128   & 0.150   \\
breastw          & 0.890 & 0.989 & 0.982 & 0.284          & 0.954 & 0.956   & 0.932 & 0.955 & 0.297 & 0.962 & 0.897 & 0.946 & 0.660 & 0.482    & 0.776 & 0.826 & 0.635 & 0.908      & 0.537 & 0.921  & 0.770  & 0.715 & 0.982 & 0.966   & 0.944   \\
campaign         & 0.287 & 0.368 & 0.354 & 0.145          & 0.352 & 0.279   & 0.289 & 0.131 & 0.158 & 0.325 & 0.283 & 0.284 & 0.163 & 0.149    & 0.113 & 0.105 & 0.267 & 0.191      & 0.299 & 0.281  & 0.237  & 0.321 & 0.315 & 0.333   & 0.247   \\
cardio           & 0.482 & 0.576 & 0.567 & 0.161          & 0.458 & 0.559   & 0.402 & 0.428 & 0.159 & 0.364 & 0.536 & 0.609 & 0.193 & 0.177    & 0.272 & 0.540 & 0.108 & 0.471      & 0.278 & 0.376  & 0.184  & 0.268 & 0.526 & 0.386   & 0.521   \\
cardiotocography & 0.335 & 0.403 & 0.502 & 0.276          & 0.361 & 0.436   & 0.324 & 0.463 & 0.272 & 0.311 & 0.408 & 0.462 & 0.271 & 0.252    & 0.258 & 0.403 & 0.188 & 0.348      & 0.338 & 0.312  & 0.250  & 0.276 & 0.422 & 0.337   & 0.419   \\
celeba           & 0.069 & 0.093 & 0.095 & 0.024          & 0.090 & 0.063   & 0.061 & 0.047 & 0.018 & 0.092 & 0.103 & 0.112 & 0.044 & 0.031    & 0.047 & 0.021 & 0.045 & 0.066      & 0.093 & 0.052  & 0.058  & 0.077 & 0.123 & 0.061   & 0.092   \\
census           & 0.088 & 0.062 & 0.062 & 0.061          & 0.073 & 0.073   & 0.088 & 0.065 & 0.069 & 0.153 & 0.085 & 0.087 & 0.062 & 0.075    & 0.058 & 0.072 & 0.095 & 0.074      & 0.086 & 0.090  & 0.083  & 0.081 & 0.089 & 0.098   & 0.089   \\
cover            & 0.070 & 0.068 & 0.113 & 0.019          & 0.026 & 0.052   & 0.054 & 0.090 & 0.019 & 0.016 & 0.099 & 0.075 & 0.044 & 0.048    & 0.056 & 0.005 & 0.022 & 0.010      & 0.046 & 0.048  & 0.025  & 0.021 & 0.068 & 0.178   & 0.050   \\
donors           & 0.148 & 0.209 & 0.265 & 0.120          & 0.135 & 0.124   & 0.182 & 0.255 & 0.109 & 0.141 & 0.139 & 0.166 & 0.086 & 0.112    & 0.123 & 0.040 & 0.119 & 0.241      & 0.143 & 0.188  & 0.164  & 0.140 & 0.136 & 0.140   & 0.072   \\
fault            & 0.473 & 0.313 & 0.325 & 0.396          & 0.360 & 0.395   & 0.522 & 0.337 & 0.388 & 0.334 & 0.401 & 0.332 & 0.361 & 0.375    & 0.496 & 0.381 & 0.473 & 0.329      & 0.392 & 0.532  & 0.417  & 0.422 & 0.412 & 0.512   & 0.474   \\
fraud            & 0.145 & 0.252 & 0.215 & 0.003          & 0.209 & 0.145   & 0.169 & 0.146 & 0.003 & 0.488 & 0.110 & 0.149 & 0.084 & 0.250    & 0.002 & 0.257 & 0.127 & 0.447      & 0.146 & 0.137  & 0.188  & 0.648 & 0.369 & 0.334   & 0.346   \\
glass            & 0.144 & 0.111 & 0.183 & 0.151          & 0.161 & 0.144   & 0.167 & 0.090 & 0.144 & 0.113 & 0.130 & 0.112 & 0.111 & 0.090    & 0.159 & 0.076 & 0.122 & 0.113      & 0.073 & 0.206  & 0.135  & 0.168 & 0.161 & 0.111   & 0.095   \\
hepatitis        & 0.304 & 0.389 & 0.295 & 0.225          & 0.328 & 0.243   & 0.252 & 0.275 & 0.214 & 0.363 & 0.277 & 0.339 & 0.253 & 0.170    & 0.221 & 0.291 & 0.231 & 0.317      & 0.165 & 0.238  & 0.215  & 0.257 & 0.319 & 0.259   & 0.430   \\
http             & 0.464 & 0.280 & 0.145 & 0.047          & 0.302 & 0.886   & 0.010 & 0.004 & 0.050 & 0.865 & 0.356 & 0.500 & 0.368 & 0.093    & 0.004 & 0.441 & 0.091 & 0.363      & 0.642 & 0.024  & 0.295  & 0.440 & 0.259 & 0.222   & 0.294   \\
internetads      & 0.297 & 0.505 & 0.505 & 0.182          & 0.523 & 0.486   & 0.296 & 0.242 & 0.232 & 0.344 & 0.291 & 0.276 & 0.207 & 0.252    & 0.197 & 0.288 & 0.237 & 0.262      & 0.295 & 0.290  & 0.275  & 0.302 & 0.281 & 0.500   & 0.354   \\
ionosphere       & 0.881 & 0.663 & 0.633 & 0.821          & 0.353 & 0.779   & 0.911 & 0.741 & 0.807 & 0.947 & 0.829 & 0.721 & 0.473 & 0.392    & 0.728 & 0.781 & 0.472 & 0.824      & 0.633 & 0.920  & 0.610  & 0.880 & 0.700 & 0.911   & 0.814   \\
landsat          & 0.212 & 0.176 & 0.164 & 0.246          & 0.231 & 0.194   & 0.258 & 0.183 & 0.250 & 0.253 & 0.175 & 0.163 & 0.230 & 0.362    & 0.272 & 0.198 & 0.329 & 0.187      & 0.200 & 0.255  & 0.203  & 0.223 & 0.181 & 0.241   & 0.210   \\
letter           & 0.166 & 0.068 & 0.077 & 0.445          & 0.078 & 0.086   & 0.203 & 0.083 & 0.433 & 0.174 & 0.113 & 0.076 & 0.083 & 0.099    & 0.252 & 0.099 & 0.208 & 0.153      & 0.367 & 0.255  & 0.181  & 0.257 & 0.088 & 0.395   & 0.130   \\
lymphography     & 0.915 & 0.907 & 0.894 & 0.090          & 0.919 & 0.972   & 0.894 & 0.491 & 0.135 & 0.767 & 0.885 & 0.935 & 0.454 & 0.254    & 0.463 & 0.897 & 0.264 & 0.417      & 0.731 & 0.805  & 0.388  & 0.381 & 1.000 & 0.705   & 0.780   \\
magic.gamma      & 0.666 & 0.588 & 0.533 & 0.539          & 0.617 & 0.638   & 0.724 & 0.579 & 0.520 & 0.632 & 0.625 & 0.589 & 0.450 & 0.499    & 0.627 & 0.326 & 0.548 & 0.692      & 0.651 & 0.730  & 0.657  & 0.664 & 0.650 & 0.767   & 0.685   \\
mammography      & 0.140 & 0.430 & 0.435 & 0.070          & 0.132 & 0.218   & 0.181 & 0.218 & 0.085 & 0.036 & 0.187 & 0.204 & 0.111 & 0.025    & 0.114 & 0.046 & 0.046 & 0.074      & 0.099 & 0.175  & 0.082  & 0.170 & 0.086 & 0.163   & 0.068   \\
musk             & 1.000 & 0.369 & 0.475 & 0.140          & 0.999 & 0.945   & 0.708 & 0.842 & 0.118 & 0.992 & 1.000 & 1.000 & 0.500 & 0.107    & 0.196 & 1.000 & 0.128 & 0.391      & 0.984 & 0.434  & 0.137  & 0.553 & 1.000 & 1.000   & 1.000   \\
optdigits        & 0.059 & 0.029 & 0.029 & 0.036          & 0.192 & 0.046   & 0.022 & 0.029 & 0.035 & 0.022 & 0.027 & 0.027 & 0.026 & 0.039    & 0.032 & 0.039 & 0.030 & 0.027      & 0.022 & 0.021  & 0.028  & 0.028 & 0.029 & 0.029   & 0.039   \\
pageblocks       & 0.547 & 0.370 & 0.520 & 0.341          & 0.319 & 0.464   & 0.556 & 0.410 & 0.292 & 0.617 & 0.531 & 0.525 & 0.255 & 0.288    & 0.632 & 0.373 & 0.285 & 0.538      & 0.493 & 0.530  & 0.507  & 0.555 & 0.503 & 0.467   & 0.554   \\
pendigits        & 0.192 & 0.177 & 0.270 & 0.048          & 0.247 & 0.260   & 0.100 & 0.186 & 0.040 & 0.069 & 0.226 & 0.219 & 0.056 & 0.022    & 0.027 & 0.075 & 0.045 & 0.060      & 0.056 & 0.089  & 0.044  & 0.044 & 0.236 & 0.116   & 0.188   \\
pima             & 0.484 & 0.536 & 0.484 & 0.412          & 0.577 & 0.510   & 0.530 & 0.404 & 0.406 & 0.498 & 0.477 & 0.492 & 0.372 & 0.366    & 0.413 & 0.476 & 0.385 & 0.476      & 0.400 & 0.528  & 0.437  & 0.447 & 0.485 & 0.500   & 0.500   \\
satellite        & 0.656 & 0.570 & 0.526 & 0.378          & 0.688 & 0.649   & 0.582 & 0.613 & 0.381 & 0.768 & 0.654 & 0.606 & 0.527 & 0.406    & 0.465 & 0.658 & 0.451 & 0.596      & 0.662 & 0.563  & 0.380  & 0.529 & 0.668 & 0.531   & 0.696   \\
satimage-2       & 0.972 & 0.797 & 0.666 & 0.042          & 0.760 & 0.918   & 0.690 & 0.857 & 0.041 & 0.682 & 0.965 & 0.872 & 0.289 & 0.052    & 0.076 & 0.949 & 0.102 & 0.484      & 0.783 & 0.507  & 0.095  & 0.138 & 0.953 & 0.836   & 0.931   \\
shuttle          & 0.184 & 0.962 & 0.905 & 0.081          & 0.965 & 0.976   & 0.193 & 0.168 & 0.109 & 0.841 & 0.907 & 0.913 & 0.438 & 0.149    & 0.072 & 0.136 & 0.135 & 0.346      & 0.779 & 0.187  & 0.247  & 0.626 & 0.948 & 0.953   & 0.958   \\
skin             & 0.289 & 0.179 & 0.183 & 0.207          & 0.232 & 0.254   & 0.290 & 0.180 & 0.221 & 0.490 & 0.220 & 0.172 & 0.226 & 0.221    & 0.285 & 0.232 & 0.173 & 0.335      & 0.175 & 0.290  & 0.316  & 0.302 & 0.235 & 0.371   & 0.455   \\
smtp             & 0.403 & 0.005 & 0.589 & 0.001          & 0.005 & 0.005   & 0.415 & 0.312 & 0.022 & 0.006 & 0.383 & 0.382 & 0.179 & 0.240    & 0.000 & 0.358 & 0.004 & 0.006      & 0.502 & 0.411  & 0.012  & 0.422 & 0.333 & 0.358   & 0.057   \\
spambase         & 0.402 & 0.544 & 0.518 & 0.344          & 0.518 & 0.488   & 0.415 & 0.387 & 0.360 & 0.349 & 0.402 & 0.409 & 0.389 & 0.456    & 0.383 & 0.387 & 0.370 & 0.433      & 0.384 & 0.407  & 0.399  & 0.400 & 0.408 & 0.485   & 0.388   \\
speech           & 0.019 & 0.019 & 0.020 & 0.022          & 0.023 & 0.021   & 0.019 & 0.016 & 0.022 & 0.019 & 0.019 & 0.018 & 0.022 & 0.018    & 0.020 & 0.019 & 0.020 & 0.018      & 0.020 & 0.019  & 0.019  & 0.020 & 0.020 & 0.018   & 0.017   \\
stamps           & 0.211 & 0.398 & 0.324 & 0.143          & 0.332 & 0.347   & 0.317 & 0.280 & 0.153 & 0.257 & 0.318 & 0.364 & 0.198 & 0.099    & 0.241 & 0.285 & 0.117 & 0.284      & 0.143 & 0.273  & 0.235  & 0.226 & 0.355 & 0.196   & 0.317   \\
thyroid          & 0.272 & 0.179 & 0.472 & 0.069          & 0.501 & 0.562   & 0.392 & 0.189 & 0.077 & 0.702 & 0.329 & 0.356 & 0.126 & 0.024    & 0.338 & 0.318 & 0.066 & 0.734      & 0.325 & 0.360  & 0.118  & 0.705 & 0.244 & 0.452   & 0.225   \\
vertebral        & 0.123 & 0.085 & 0.110 & 0.124          & 0.091 & 0.097   & 0.095 & 0.089 & 0.130 & 0.101 & 0.107 & 0.099 & 0.134 & 0.107    & 0.118 & 0.124 & 0.115 & 0.111      & 0.150 & 0.098  & 0.133  & 0.119 & 0.153 & 0.090   & 0.103   \\
vowels           & 0.166 & 0.034 & 0.083 & 0.314          & 0.078 & 0.162   & 0.443 & 0.127 & 0.326 & 0.085 & 0.196 & 0.069 & 0.041 & 0.037    & 0.178 & 0.154 & 0.219 & 0.295      & 0.311 & 0.504  & 0.166  & 0.417 & 0.255 & 0.521   & 0.210   \\
waveform         & 0.122 & 0.057 & 0.040 & 0.078          & 0.048 & 0.056   & 0.133 & 0.040 & 0.071 & 0.040 & 0.052 & 0.044 & 0.032 & 0.061    & 0.150 & 0.042 & 0.063 & 0.150      & 0.050 & 0.109  & 0.037  & 0.043 & 0.060 & 0.061   & 0.101   \\
wbc              & 0.691 & 0.883 & 0.882 & 0.037          & 0.728 & 0.948   & 0.743 & 0.898 & 0.077 & 0.839 & 0.813 & 0.913 & 0.327 & 0.069    & 0.358 & 0.736 & 0.211 & 0.431      & 0.758 & 0.722  & 0.348  & 0.194 & 1.000 & 0.679   & 0.902   \\
wdbc             & 0.688 & 0.760 & 0.493 & 0.155          & 0.761 & 0.702   & 0.521 & 0.527 & 0.128 & 0.395 & 0.539 & 0.613 & 0.152 & 0.063    & 0.039 & 0.589 & 0.065 & 0.568      & 0.483 & 0.465  & 0.074  & 0.157 & 0.393 & 0.290   & 0.568   \\
wilt             & 0.040 & 0.037 & 0.042 & 0.081          & 0.039 & 0.044   & 0.049 & 0.036 & 0.083 & 0.153 & 0.035 & 0.032 & 0.047 & 0.046    & 0.041 & 0.065 & 0.109 & 0.115      & 0.076 & 0.054  & 0.211  & 0.163 & 0.036 & 0.071   & 0.037   \\
wine             & 0.170 & 0.364 & 0.195 & 0.061          & 0.412 & 0.207   & 0.081 & 0.250 & 0.064 & 0.737 & 0.135 & 0.264 & 0.120 & 0.116    & 0.126 & 0.229 & 0.087 & 0.086      & 0.075 & 0.074  & 0.064  & 0.103 & 0.335 & 0.219   & 0.385   \\
wpbc             & 0.227 & 0.234 & 0.217 & 0.206          & 0.241 & 0.237   & 0.234 & 0.227 & 0.210 & 0.257 & 0.222 & 0.229 & 0.214 & 0.240    & 0.234 & 0.214 & 0.234 & 0.236      & 0.238 & 0.227  & 0.238  & 0.231 & 0.255 & 0.269   & 0.224   \\
yeast            & 0.314 & 0.308 & 0.332 & 0.326          & 0.328 & 0.304   & 0.294 & 0.330 & 0.315 & 0.298 & 0.303 & 0.302 & 0.353 & 0.350    & 0.284 & 0.332 & 0.318 & 0.309      & 0.320 & 0.295  & 0.306  & 0.306 & 0.292 & 0.305   & 0.299   \\
CIFAR10          & 0.103 & 0.065 & 0.067 & 0.115          & 0.075 & 0.089   & 0.102 & 0.086 & 0.115 & 0.084 & 0.102 & 0.101 & 0.062 & 0.073    & 0.060 & 0.102 & 0.070 & 0.085      & 0.102 & 0.104  & 0.078  & 0.092 & 0.557 & 0.499   & 0.531   \\
MNIST-C          & 0.173 & 0.050 & 0.050 & 0.128          & 0.126 & 0.178   & 0.191 & 0.101 & 0.127 & 0.166 & 0.179 & 0.170 & 0.092 & 0.097    & 0.096 & 0.177 & 0.098 & 0.154      & 0.178 & 0.192  & 0.141  & 0.157 & 0.281 & 0.281   & 0.305   \\
MVTec-AD         & 0.570 & 0.236 & 0.236 & 0.536          & 0.546 & 0.570   & 0.580 & 0.464 & 0.532 & 0.451 & 0.555 & 0.540 & 0.362 & 0.387    & 0.317 & 0.546 & 0.404 & 0.454      & 0.546 & 0.578  & 0.439  & 0.517 & 0.738 & 0.444   & 0.560   \\
SVHN             & 0.079 & 0.050 & 0.050 & 0.084          & 0.064 & 0.073   & 0.079 & 0.064 & 0.083 & 0.068 & 0.078 & 0.078 & 0.059 & 0.063    & 0.060 & 0.078 & 0.068 & 0.074      & 0.078 & 0.080  & 0.069  & 0.077 & 0.060 & 0.058   & 0.065   \\
mnist            & 0.386 & 0.092 & 0.092 & 0.241          & 0.109 & 0.290   & 0.409 & 0.170 & 0.233 & 0.308 & 0.385 & 0.381 & 0.215 & 0.253    & 0.237 & 0.297 & 0.232 & 0.259      & 0.374 & 0.400  & 0.276  & 0.368 & 0.980 & 0.238   & 0.581   \\
FashionMNIST     & 0.329 & 0.050 & 0.050 & 0.194          & 0.269 & 0.320   & 0.346 & 0.180 & 0.188 & 0.245 & 0.329 & 0.319 & 0.138 & 0.181    & 0.106 & 0.328 & 0.158 & 0.297      & 0.325 & 0.339  & 0.213  & 0.267 & 0.987 & 0.321   & 0.648   \\
20news           & 0.067 & 0.061 & 0.062 & 0.087          & 0.061 & 0.062   & 0.069 & 0.062 & 0.088 & 0.072 & 0.064 & 0.062 & 0.054 & 0.058    & 0.055 & 0.063 & 0.063 & 0.056      & 0.063 & 0.072  & 0.060  & 0.068 & 0.110 & 0.088   & 0.116   \\
agnews           & 0.072 & 0.059 & 0.058 & 0.125          & 0.059 & 0.064   & 0.082 & 0.064 & 0.125 & 0.077 & 0.068 & 0.061 & 0.053 & 0.053    & 0.051 & 0.066 & 0.069 & 0.050      & 0.062 & 0.085  & 0.063  & 0.076 & 0.149 & 0.106   & 0.276   \\
amazon           & 0.061 & 0.060 & 0.055 & 0.058          & 0.059 & 0.058   & 0.062 & 0.054 & 0.058 & 0.062 & 0.059 & 0.057 & 0.049 & 0.046    & 0.050 & 0.058 & 0.052 & 0.050      & 0.057 & 0.062  & 0.055  & 0.057 & 0.051 & 0.051   & 0.054   \\
imdb             & 0.047 & 0.050 & 0.045 & 0.049          & 0.047 & 0.047   & 0.047 & 0.046 & 0.049 & 0.049 & 0.047 & 0.046 & 0.049 & 0.053    & 0.050 & 0.047 & 0.054 & 0.051      & 0.046 & 0.047  & 0.047  & 0.047 & 0.053 & 0.050   & 0.056   \\
yelp             & 0.073 & 0.072 & 0.065 & 0.085          & 0.070 & 0.070   & 0.083 & 0.067 & 0.085 & 0.075 & 0.073 & 0.069 & 0.049 & 0.058    & 0.051 & 0.068 & 0.054 & 0.056      & 0.069 & 0.085  & 0.054  & 0.066 & 0.054 & 0.054   & 0.057   \\
\midrule
\textbf{Average} & 0.318 & 0.288 & 0.296 & 0.179          & 0.308 & 0.336   & 0.308 & 0.260 & 0.181 & 0.337 & 0.324 & 0.328 & 0.198 & 0.170    & 0.199 & 0.285 & 0.185 & 0.283      & 0.301 & 0.295  & 0.216  & 0.288 & 0.368 & 0.336   & 0.348 
\\ 
\bottomrule
\end{tabular}
\caption{PR AUC for the unsupervised setting on ADBench}
\label{tab:adbench2}
\end{sidewaystable}

\begin{sidewaystable}[h!]
\renewcommand\thetable{B.4}
\centering
\setlength{\tabcolsep}{0.7mm} 
\fontsize{6pt}{7pt}\selectfont 
\begin{tabular}{lcccccccccccccccccccccccc}
\toprule
    & CBLOF & COPOD & ECOD  & FeatureBagging & HBOS  & IForest & kNN   & LODA  & LOF   & MCD   & OCSVM & PCA   & DAGMM & DeepSVDD & DROCC & GOAD  & ICL   & PlanarFlow & DDPM  & DTE-NP & DTE-IG & DTE-C & ALTBI   \\ 
\midrule
aloi             & 0.537 & 0.495 & 0.517 & 0.491          & 0.522 & 0.507   & 0.510 & 0.492 & 0.488 & 0.485 & 0.543 & 0.540 & 0.508 & 0.509    & 0.500 & 0.480 & 0.475 & 0.485      & 0.499 & 0.512  & 0.509  & 0.504 & 0.539   \\
annthyroid       & 0.901 & 0.768 & 0.785 & 0.890          & 0.660 & 0.903   & 0.928 & 0.774 & 0.886 & 0.902 & 0.885 & 0.852 & 0.722 & 0.550    & 0.889 & 0.810 & 0.811 & 0.932      & 0.888 & 0.929  & 0.876  & 0.975 & 0.655   \\
backdoor         & 0.697 & 0.500 & 0.500 & 0.948          & 0.708 & 0.749   & 0.938 & 0.476 & 0.953 & 0.851 & 0.625 & 0.646 & 0.544 & 0.911    & 0.943 & 0.529 & 0.936 & 0.760      & 0.809 & 0.933  & 0.940  & 0.917 & 0.927   \\
breastw          & 0.991 & 0.995 & 0.991 & 0.591          & 0.992 & 0.995   & 0.991 & 0.981 & 0.889 & 0.987 & 0.994 & 0.992 & 0.895 & 0.970    & 0.473 & 0.989 & 0.983 & 0.979      & 0.987 & 0.993  & 0.787  & 0.928 & 0.982   \\
campaign         & 0.771 & 0.782 & 0.769 & 0.691          & 0.771 & 0.736   & 0.785 & 0.589 & 0.706 & 0.785 & 0.777 & 0.771 & 0.615 & 0.622    & 0.500 & 0.479 & 0.809 & 0.698      & 0.745 & 0.788  & 0.748  & 0.780 & 0.749   \\
cardio           & 0.935 & 0.932 & 0.950 & 0.921          & 0.807 & 0.933   & 0.920 & 0.913 & 0.922 & 0.828 & 0.956 & 0.965 & 0.779 & 0.654    & 0.621 & 0.960 & 0.800 & 0.889      & 0.869 & 0.918  & 0.738  & 0.873 & 0.869   \\
cardiotocography & 0.676 & 0.664 & 0.793 & 0.636          & 0.612 & 0.742   & 0.621 & 0.728 & 0.645 & 0.571 & 0.752 & 0.789 & 0.671 & 0.478    & 0.460 & 0.761 & 0.542 & 0.699      & 0.545 & 0.638  & 0.524  & 0.601 & 0.686   \\
celeba           & 0.793 & 0.757 & 0.763 & 0.469          & 0.767 & 0.712   & 0.731 & 0.625 & 0.437 & 0.844 & 0.798 & 0.805 & 0.638 & 0.562    & 0.689 & 0.438 & 0.722 & 0.716      & 0.786 & 0.704  & 0.745  & 0.822 & 0.701   \\
census           & 0.708 & 0.500 & 0.500 & 0.559          & 0.625 & 0.626   & 0.723 & 0.511 & 0.585 & 0.741 & 0.700 & 0.705 & 0.522 & 0.542    & 0.554 & 0.352 & 0.706 & 0.593      & 0.702 & 0.721  & 0.618  & 0.696 & 0.704   \\
cover            & 0.940 & 0.882 & 0.919 & 0.992          & 0.711 & 0.863   & 0.975 & 0.949 & 0.992 & 0.700 & 0.962 & 0.944 & 0.759 & 0.491    & 0.958 & 0.138 & 0.893 & 0.475      & 0.984 & 0.977  & 0.958  & 0.978 & 0.947   \\
donors           & 0.935 & 0.815 & 0.887 & 0.952          & 0.812 & 0.894   & 0.995 & 0.635 & 0.970 & 0.819 & 0.921 & 0.881 & 0.622 & 0.730    & 0.742 & 0.336 & 0.999 & 0.916      & 0.825 & 0.993  & 0.993  & 0.982 & 0.960   \\
fault            & 0.590 & 0.491 & 0.504 & 0.483          & 0.531 & 0.559   & 0.587 & 0.503 & 0.474 & 0.594 & 0.572 & 0.559 & 0.529 & 0.543    & 0.557 & 0.589 & 0.606 & 0.575      & 0.611 & 0.586  & 0.594  & 0.595 & 0.734   \\
fraud            & 0.949 & 0.943 & 0.949 & 0.948          & 0.950 & 0.947   & 0.954 & 0.891 & 0.944 & 0.911 & 0.956 & 0.954 & 0.853 & 0.831    & 0.500 & 0.698 & 0.928 & 0.907      & 0.937 & 0.956  & 0.908  & 0.935 & 0.954   \\
glass            & 0.894 & 0.760 & 0.711 & 0.885          & 0.826 & 0.811   & 0.920 & 0.673 & 0.888 & 0.797 & 0.697 & 0.734 & 0.653 & 0.837    & 0.649 & 0.590 & 0.994 & 0.853      & 0.667 & 0.896  & 0.985  & 0.924 & 0.741   \\
hepatitis        & 0.863 & 0.809 & 0.738 & 0.678          & 0.848 & 0.827   & 0.965 & 0.690 & 0.669 & 0.806 & 0.906 & 0.845 & 0.702 & 0.996    & 0.518 & 0.845 & 0.999 & 0.958      & 0.977 & 0.932  & 0.999  & 0.988 & 0.593   \\
http             & 0.999 & 0.992 & 0.980 & 0.921          & 0.986 & 0.994   & 1.000 & 0.477 & 1.000 & 1.000 & 1.000 & 1.000 & 0.918 & 0.613    & 0.500 & 0.997 & 0.982 & 0.994      & 1.000 & 1.000  & 0.807  & 0.995 & 0.994   \\
internetads      & 0.652 & 0.659 & 0.660 & 0.714          & 0.492 & 0.479   & 0.681 & 0.587 & 0.717 & 0.477 & 0.656 & 0.651 & 0.495 & 0.730    & 0.534 & 0.657 & 0.722 & 0.709      & 0.658 & 0.700  & 0.715  & 0.776 & 0.909   \\
ionosphere       & 0.968 & 0.783 & 0.718 & 0.945          & 0.707 & 0.912   & 0.974 & 0.856 & 0.943 & 0.954 & 0.963 & 0.891 & 0.740 & 0.972    & 0.611 & 0.915 & 0.990 & 0.969      & 0.946 & 0.978  & 0.952  & 0.954 & 0.934   \\
landsat          & 0.572 & 0.493 & 0.420 & 0.664          & 0.732 & 0.588   & 0.683 & 0.447 & 0.666 & 0.568 & 0.480 & 0.439 & 0.563 & 0.594    & 0.539 & 0.405 & 0.651 & 0.509      & 0.514 & 0.682  & 0.447  & 0.528 & 0.634   \\
letter           & 0.332 & 0.365 & 0.454 & 0.448          & 0.359 & 0.320   & 0.354 & 0.302 & 0.448 & 0.315 & 0.322 & 0.303 & 0.390 & 0.364    & 0.553 & 0.311 & 0.427 & 0.387      & 0.381 & 0.344  & 0.399  & 0.367 & 0.829   \\
lymphography     & 0.998 & 0.995 & 0.995 & 0.966          & 0.997 & 0.995   & 0.999 & 0.670 & 0.982 & 0.989 & 1.000 & 0.999 & 0.949 & 0.997    & 0.324 & 0.999 & 1.000 & 0.996      & 0.999 & 0.999  & 1.000  & 0.990 & 0.994   \\
magic.gamma      & 0.758 & 0.680 & 0.636 & 0.842          & 0.745 & 0.771   & 0.833 & 0.705 & 0.834 & 0.737 & 0.743 & 0.706 & 0.592 & 0.630    & 0.788 & 0.695 & 0.756 & 0.741      & 0.860 & 0.836  & 0.865  & 0.875 & 0.836   \\
mammography      & 0.847 & 0.906 & 0.907 & 0.863          & 0.850 & 0.880   & 0.876 & 0.896 & 0.855 & 0.729 & 0.886 & 0.899 & 0.760 & 0.715    & 0.818 & 0.699 & 0.719 & 0.789      & 0.810 & 0.876  & 0.846  & 0.864 & 0.857   \\
musk             & 1.000 & 0.997 & 0.999 & 1.000          & 1.000 & 0.906   & 1.000 & 0.997 & 1.000 & 0.939 & 1.000 & 1.000 & 0.950 & 1.000    & 0.330 & 1.000 & 0.994 & 0.767      & 1.000 & 1.000  & 0.942  & 1.000 & 1.000   \\
optdigits        & 0.835 & 0.500 & 0.500 & 0.963          & 0.899 & 0.811   & 0.937 & 0.328 & 0.967 & 0.649 & 0.634 & 0.582 & 0.400 & 0.395    & 0.853 & 0.675 & 0.972 & 0.341      & 0.908 & 0.943  & 0.798  & 0.824 & 0.871   \\
pageblocks       & 0.912 & 0.809 & 0.880 & 0.911          & 0.656 & 0.826   & 0.897 & 0.836 & 0.913 & 0.871 & 0.886 & 0.861 & 0.828 & 0.784    & 0.923 & 0.881 & 0.884 & 0.849      & 0.869 & 0.893  & 0.857  & 0.899 & 0.926   \\
pendigits        & 0.967 & 0.907 & 0.930 & 0.995          & 0.936 & 0.972   & 0.999 & 0.921 & 0.991 & 0.837 & 0.964 & 0.944 & 0.565 & 0.463    & 0.759 & 0.900 & 0.967 & 0.835      & 0.981 & 0.996  & 0.970  & 0.978 & 0.961   \\
pima             & 0.729 & 0.666 & 0.606 & 0.719          & 0.748 & 0.743   & 0.769 & 0.627 & 0.705 & 0.736 & 0.715 & 0.723 & 0.545 & 0.580    & 0.475 & 0.623 & 0.797 & 0.722      & 0.703 & 0.815  & 0.686  & 0.699 & 0.727   \\
satellite        & 0.732 & 0.683 & 0.622 & 0.801          & 0.855 & 0.775   & 0.822 & 0.697 & 0.803 & 0.728 & 0.739 & 0.666 & 0.728 & 0.762    & 0.734 & 0.688 & 0.852 & 0.723      & 0.777 & 0.821  & 0.765  & 0.786 & 0.797   \\
satimage-2       & 0.994 & 0.979 & 0.971 & 0.995          & 0.980 & 0.991   & 0.997 & 0.987 & 0.994 & 0.999 & 0.996 & 0.982 & 0.918 & 0.929    & 0.992 & 0.990 & 0.995 & 0.967      & 0.996 & 0.997  & 0.953  & 0.993 & 0.998   \\
shuttle          & 0.997 & 0.995 & 0.993 & 0.869          & 0.986 & 0.997   & 0.999 & 0.717 & 1.000 & 0.990 & 0.996 & 0.994 & 0.846 & 0.998    & 0.500 & 0.704 & 0.999 & 0.865      & 0.999 & 0.999  & 0.999  & 0.998 & 0.886   \\
skin             & 0.918 & 0.472 & 0.491 & 0.784          & 0.769 & 0.894   & 0.995 & 0.755 & 0.863 & 0.884 & 0.903 & 0.597 & 0.679 & 0.600    & 0.895 & 0.650 & 0.066 & 0.913      & 0.887 & 0.989  & 0.987  & 0.918 & 0.925   \\
smtp             & 0.873 & 0.912 & 0.883 & 0.848          & 0.828 & 0.904   & 0.924 & 0.730 & 0.934 & 0.949 & 0.847 & 0.818 & 0.871 & 0.852    & 0.571 & 0.788 & 0.744 & 0.842      & 0.954 & 0.930  & 0.816  & 0.953 & 0.916   \\
spambase         & 0.815 & 0.721 & 0.688 & 0.696          & 0.779 & 0.852   & 0.834 & 0.724 & 0.732 & 0.807 & 0.817 & 0.814 & 0.694 & 0.702    & 0.754 & 0.818 & 0.835 & 0.823      & 0.645 & 0.837  & 0.775  & 0.830 & 0.663   \\
speech           & 0.359 & 0.370 & 0.360 & 0.375          & 0.367 & 0.377   & 0.364 & 0.380 & 0.375 & 0.388 & 0.366 & 0.364 & 0.507 & 0.489    & 0.490 & 0.366 & 0.489 & 0.486      & 0.370 & 0.414  & 0.396  & 0.382 & 0.421   \\
stamps           & 0.934 & 0.931 & 0.876 & 0.942          & 0.918 & 0.935   & 0.959 & 0.919 & 0.937 & 0.849 & 0.937 & 0.927 & 0.801 & 0.711    & 0.502 & 0.815 & 0.967 & 0.873      & 0.918 & 0.979  & 0.934  & 0.916 & 0.955   \\
thyroid          & 0.985 & 0.938 & 0.976 & 0.932          & 0.987 & 0.990   & 0.987 & 0.961 & 0.927 & 0.985 & 0.986 & 0.986 & 0.911 & 0.888    & 0.950 & 0.952 & 0.954 & 0.984      & 0.980 & 0.986  & 0.894  & 0.987 & 0.947   \\
vertebral        & 0.544 & 0.263 & 0.420 & 0.641          & 0.401 & 0.456   & 0.577 & 0.317 & 0.643 & 0.471 & 0.505 & 0.421 & 0.506 & 0.448    & 0.438 & 0.467 & 0.792 & 0.498      & 0.707 & 0.543  & 0.746  & 0.664 & 0.285   \\
vowels           & 0.787 & 0.528 & 0.615 & 0.853          & 0.533 & 0.618   & 0.822 & 0.555 & 0.863 & 0.277 & 0.759 & 0.523 & 0.426 & 0.557    & 0.547 & 0.685 & 0.851 & 0.546      & 0.864 & 0.814  & 0.857  & 0.869 & 0.963   \\
waveform         & 0.729 & 0.724 & 0.594 & 0.770          & 0.693 & 0.723   & 0.752 & 0.610 & 0.760 & 0.584 & 0.704 & 0.647 & 0.519 & 0.599    & 0.677 & 0.650 & 0.687 & 0.648      & 0.622 & 0.745  & 0.737  & 0.652 & 0.667   \\
wbc              & 0.983 & 0.994 & 0.994 & 0.581          & 0.990 & 0.994   & 0.991 & 0.979 & 0.805 & 0.989 & 0.996 & 0.994 & 0.868 & 0.914    & 0.442 & 0.991 & 0.997 & 0.960      & 0.992 & 0.995  & 0.910  & 0.805 & 0.980   \\
wdbc             & 0.987 & 0.992 & 0.967 & 0.996          & 0.986 & 0.987   & 0.991 & 0.970 & 0.996 & 0.970 & 0.993 & 0.991 & 0.738 & 0.993    & 0.401 & 0.990 & 0.998 & 0.989      & 0.993 & 0.995  & 0.996  & 0.985 & 0.995   \\
wilt             & 0.429 & 0.321 & 0.375 & 0.734          & 0.391 & 0.480   & 0.637 & 0.411 & 0.688 & 0.817 & 0.348 & 0.261 & 0.418 & 0.344    & 0.495 & 0.514 & 0.764 & 0.746      & 0.717 & 0.629  & 0.938  & 0.851 & 0.378   \\
wine             & 0.978 & 0.864 & 0.739 & 0.979          & 0.956 & 0.939   & 0.992 & 0.909 & 0.984 & 0.973 & 0.978 & 0.938 & 0.662 & 0.922    & 0.438 & 0.941 & 0.999 & 0.954      & 0.996 & 0.994  & 1.000  & 1.000 & 0.724   \\
wpbc             & 0.596 & 0.523 & 0.495 & 0.568          & 0.609 & 0.563   & 0.637 & 0.513 & 0.574 & 0.634 & 0.534 & 0.525 & 0.470 & 0.827    & 0.438 & 0.514 & 0.966 & 0.575      & 0.665 & 0.832  & 0.707  & 0.689 & 0.491   \\
yeast            & 0.504 & 0.389 & 0.446 & 0.464          & 0.429 & 0.418   & 0.447 & 0.465 & 0.458 & 0.431 & 0.448 & 0.432 & 0.510 & 0.476    & 0.484 & 0.525 & 0.490 & 0.451      & 0.491 & 0.446  & 0.486  & 0.471 & 0.468   \\
CIFAR10          & 0.679 & 0.550 & 0.569 & 0.703          & 0.579 & 0.640   & 0.675 & 0.616 & 0.703 & 0.652 & 0.678 & 0.674 & 0.540 & 0.561    & 0.496 & 0.675 & 0.636 & 0.628      & 0.679 & 0.678  & 0.624  & 0.685 & 0.948   \\
MNIST-C          & 0.811 & 0.500 & 0.500 & 0.873          & 0.704 & 0.768   & 0.841 & 0.694 & 0.872 & 0.753 & 0.796 & 0.784 & 0.637 & 0.647    & 0.572 & 0.793 & 0.856 & 0.712      & 0.801 & 0.847  & 0.799  & 0.861 & 0.858   \\
MVTec-AD         & 0.800 & 0.500 & 0.500 & 0.805          & 0.760 & 0.774   & 0.815 & 0.723 & 0.804 & 0.868 & 0.774 & 0.764 & 0.647 & 0.896    & 0.605 & 0.771 & 0.948 & 0.728      & 0.780 & 0.897  & 0.859  & 0.894 & 0.929   \\
SVHN             & 0.610 & 0.500 & 0.500 & 0.639          & 0.547 & 0.590   & 0.617 & 0.545 & 0.638 & 0.589 & 0.613 & 0.608 & 0.534 & 0.539    & 0.500 & 0.608 & 0.617 & 0.589      & 0.614 & 0.621  & 0.592  & 0.629 & 0.604   \\
mnist            & 0.911 & 0.500 & 0.500 & 0.926          & 0.623 & 0.866   & 0.939 & 0.647 & 0.929 & 0.883 & 0.906 & 0.902 & 0.722 & 0.664    & 0.831 & 0.901 & 0.901 & 0.819      & 0.873 & 0.940  & 0.808  & 0.874 & 0.927   \\
FashionMNIST     & 0.891 & 0.500 & 0.500 & 0.917          & 0.754 & 0.842   & 0.899 & 0.793 & 0.916 & 0.844 & 0.882 & 0.876 & 0.708 & 0.755    & 0.516 & 0.880 & 0.906 & 0.822      & 0.885 & 0.901  & 0.843  & 0.902 & 0.916   \\
20news           & 0.571 & 0.529 & 0.541 & 0.603          & 0.536 & 0.549   & 0.574 & 0.535 & 0.602 & 0.629 & 0.563 & 0.544 & 0.515 & 0.556    & 0.528 & 0.549 & 0.613 & 0.516      & 0.549 & 0.600  & 0.583  & 0.643 & 0.807   \\
agnews           & 0.628 & 0.551 & 0.551 & 0.746          & 0.557 & 0.584   & 0.671 & 0.570 & 0.746 & 0.680 & 0.606 & 0.569 & 0.510 & 0.498    & 0.497 & 0.599 & 0.626 & 0.502      & 0.578 & 0.680  & 0.566  & 0.682 & 0.889   \\
amazon           & 0.582 & 0.568 & 0.538 & 0.579          & 0.563 & 0.564   & 0.606 & 0.522 & 0.579 & 0.604 & 0.565 & 0.549 & 0.505 & 0.512    & 0.500 & 0.561 & 0.542 & 0.499      & 0.551 & 0.608  & 0.519  & 0.567 & 0.585   \\
imdb             & 0.499 & 0.511 & 0.469 & 0.495          & 0.499 & 0.495   & 0.501 & 0.472 & 0.496 & 0.512 & 0.487 & 0.480 & 0.486 & 0.500    & 0.514 & 0.485 & 0.523 & 0.492      & 0.479 & 0.504  & 0.510  & 0.481 & 0.575   \\
yelp             & 0.638 & 0.602 & 0.574 & 0.671          & 0.600 & 0.611   & 0.681 & 0.563 & 0.672 & 0.662 & 0.621 & 0.592 & 0.499 & 0.499    & 0.507 & 0.611 & 0.558 & 0.536      & 0.593 & 0.687  & 0.573  & 0.599 & 0.564   \\
\midrule
\textbf{Average}          & 0.781 & 0.689 & 0.688 & 0.770          & 0.727 & 0.757   & 0.809 & 0.673 & 0.785 & 0.751 & 0.765 & 0.740 & 0.651 & 0.679    & 0.603 & 0.688 & 0.794 & 0.732      & 0.779 & 0.815  & 0.779  & 0.804 & 0.794  
\\ 
\bottomrule
\end{tabular}
\caption{ROC AUC for the semi-supervised setting on ADBench}
\label{tab:adbench3}
\end{sidewaystable}

\begin{sidewaystable}[h!]
\renewcommand\thetable{B.5}
\centering
\setlength{\tabcolsep}{0.7mm} 
\fontsize{6pt}{7pt}\selectfont 
\begin{tabular}{lccccccccccccccccccccccc}
\toprule
    & CBLOF & COPOD & ECOD  & FeatureBagging & HBOS  & IForest & kNN   & LODA  & LOF   & MCD   & OCSVM & PCA   & DAGMM & DeepSVDD & DROCC & GOAD  & ICL   & PlanarFlow & DDPM  & DTE-NP & DTE-IG & DTE-C & ALTBI   \\ 
\midrule
aloi             & 0.064 & 0.057 & 0.061 & 0.068          & 0.064 & 0.058   & 0.060 & 0.059 & 0.065 & 0.056 & 0.065 & 0.065 & 0.061 & 0.062    & 0.059 & 0.057 & 0.055 & 0.055      & 0.060 & 0.061  & 0.060  & 0.058 & 0.040   \\
annthyroid       & 0.636 & 0.296 & 0.400 & 0.485          & 0.390 & 0.590   & 0.681 & 0.490 & 0.535 & 0.597 & 0.601 & 0.566 & 0.480 & 0.278    & 0.637 & 0.587 & 0.458 & 0.652      & 0.629 & 0.682  & 0.499  & 0.829 & 0.207   \\
backdoor         & 0.091 & 0.048 & 0.048 & 0.495          & 0.086 & 0.094   & 0.465 & 0.060 & 0.535 & 0.222 & 0.077 & 0.079 & 0.075 & 0.848    & 0.846 & 0.063 & 0.892 & 0.322      & 0.142 & 0.457  & 0.820  & 0.624 & 0.815   \\
breastw          & 0.991 & 0.994 & 0.992 & 0.524          & 0.991 & 0.995   & 0.989 & 0.968 & 0.800 & 0.983 & 0.994 & 0.992 & 0.910 & 0.960    & 0.632 & 0.988 & 0.968 & 0.975      & 0.986 & 0.992  & 0.814  & 0.883 & 0.960   \\
campaign         & 0.486 & 0.511 & 0.495 & 0.333          & 0.497 & 0.457   & 0.490 & 0.298 & 0.402 & 0.479 & 0.494 & 0.488 & 0.324 & 0.370    & 0.203 & 0.231 & 0.489 & 0.428      & 0.489 & 0.500  & 0.462  & 0.469 & 0.340   \\
cardio           & 0.809 & 0.749 & 0.786 & 0.716          & 0.589 & 0.786   & 0.772 & 0.725 & 0.702 & 0.671 & 0.836 & 0.862 & 0.559 & 0.389    & 0.512 & 0.848 & 0.479 & 0.689      & 0.693 & 0.774  & 0.411  & 0.693 & 0.530   \\
cardiotocography & 0.617 & 0.561 & 0.690 & 0.570          & 0.507 & 0.629   & 0.574 & 0.606 & 0.573 & 0.528 & 0.662 & 0.697 & 0.597 & 0.458    & 0.439 & 0.675 & 0.487 & 0.593      & 0.513 & 0.587  & 0.396  & 0.533 & 0.490   \\
celeba           & 0.185 & 0.165 & 0.169 & 0.039          & 0.168 & 0.117   & 0.119 & 0.095 & 0.036 & 0.190 & 0.203 & 0.210 & 0.090 & 0.071    & 0.077 & 0.040 & 0.097 & 0.129      & 0.180 & 0.107  & 0.134  & 0.142 & 0.042   \\
census           & 0.203 & 0.117 & 0.117 & 0.120          & 0.140 & 0.142   & 0.217 & 0.134 & 0.137 & 0.290 & 0.203 & 0.200 & 0.132 & 0.154    & 0.143 & 0.087 & 0.212 & 0.147      & 0.197 & 0.211  & 0.163  & 0.179 & 0.106   \\
cover            & 0.160 & 0.123 & 0.192 & 0.781          & 0.054 & 0.087   & 0.558 & 0.226 & 0.829 & 0.031 & 0.223 & 0.162 & 0.098 & 0.027    & 0.313 & 0.011 & 0.345 & 0.020      & 0.733 & 0.600  & 0.804  & 0.637 & 0.227   \\
donors           & 0.465 & 0.335 & 0.413 & 0.653          & 0.363 & 0.405   & 0.891 & 0.254 & 0.634 & 0.312 & 0.427 & 0.352 & 0.195 & 0.428    & 0.302 & 0.090 & 0.984 & 0.493      & 0.267 & 0.856  & 0.958  & 0.713 & 0.433   \\
fault            & 0.613 & 0.532 & 0.517 & 0.508          & 0.539 & 0.592   & 0.620 & 0.545 & 0.504 & 0.634 & 0.611 & 0.604 & 0.568 & 0.555    & 0.578 & 0.621 & 0.632 & 0.604      & 0.648 & 0.622  & 0.638  & 0.639 & 0.560   \\
fraud            & 0.278 & 0.384 & 0.332 & 0.631          & 0.323 & 0.182   & 0.387 & 0.366 & 0.551 & 0.601 & 0.296 & 0.269 & 0.156 & 0.483    & 0.003 & 0.294 & 0.539 & 0.628      & 0.692 & 0.421  & 0.511  & 0.621 & 0.354   \\
glass            & 0.317 & 0.201 & 0.250 & 0.361          & 0.276 & 0.214   & 0.423 & 0.156 & 0.381 & 0.203 & 0.268 & 0.210 & 0.186 & 0.524    & 0.231 & 0.183 & 0.924 & 0.309      & 0.312 & 0.374  & 0.806  & 0.415 & 0.362   \\
hepatitis        & 0.634 & 0.561 & 0.458 & 0.446          & 0.635 & 0.554   & 0.903 & 0.502 & 0.437 & 0.568 & 0.776 & 0.649 & 0.544 & 0.987    & 0.349 & 0.658 & 0.998 & 0.896      & 0.951 & 0.823  & 0.998  & 0.958 & 0.239   \\
http             & 0.903 & 0.463 & 0.252 & 0.082          & 0.390 & 0.534   & 1.000 & 0.075 & 0.971 & 0.922 & 0.999 & 0.917 & 0.575 & 0.361    & 0.007 & 0.684 & 0.708 & 0.522      & 1.000 & 0.971  & 0.788  & 0.555 & 0.233   \\
internetads      & 0.470 & 0.617 & 0.619 & 0.493          & 0.308 & 0.292   & 0.492 & 0.393 & 0.504 & 0.344 & 0.482 & 0.470 & 0.318 & 0.516    & 0.431 & 0.474 & 0.600 & 0.476      & 0.477 & 0.513  & 0.587  & 0.552 & 0.820   \\
ionosphere       & 0.973 & 0.785 & 0.756 & 0.949          & 0.646 & 0.917   & 0.980 & 0.852 & 0.946 & 0.967 & 0.975 & 0.909 & 0.775 & 0.981    & 0.717 & 0.932 & 0.991 & 0.976      & 0.964 & 0.982  & 0.969  & 0.968 & 0.910   \\
landsat          & 0.369 & 0.338 & 0.311 & 0.615          & 0.601 & 0.473   & 0.549 & 0.357 & 0.614 & 0.397 & 0.370 & 0.327 & 0.403 & 0.494    & 0.376 & 0.312 & 0.531 & 0.342      & 0.348 & 0.545  & 0.327  & 0.368 & 0.276   \\
letter           & 0.083 & 0.089 & 0.107 & 0.117          & 0.087 & 0.082   & 0.087 & 0.080 & 0.113 & 0.081 & 0.083 & 0.080 & 0.104 & 0.089    & 0.157 & 0.081 & 0.128 & 0.092      & 0.095 & 0.086  & 0.102  & 0.090 & 0.200   \\
lymphography     & 0.983 & 0.939 & 0.944 & 0.727          & 0.966 & 0.944   & 0.992 & 0.241 & 0.842 & 0.868 & 1.000 & 0.985 & 0.735 & 0.968    & 0.309 & 0.988 & 1.000 & 0.962      & 0.993 & 0.993  & 0.998  & 0.868 & 0.917   \\
magic.gamma      & 0.802 & 0.722 & 0.679 & 0.869          & 0.772 & 0.803   & 0.859 & 0.758 & 0.864 & 0.772 & 0.792 & 0.752 & 0.645 & 0.695    & 0.832 & 0.761 & 0.813 & 0.785      & 0.880 & 0.862  & 0.887  & 0.897 & 0.771   \\
mammography      & 0.411 & 0.546 & 0.552 & 0.293          & 0.213 & 0.379   & 0.413 & 0.432 & 0.341 & 0.080 & 0.405 & 0.417 & 0.220 & 0.275    & 0.272 & 0.278 & 0.171 & 0.185      & 0.199 & 0.421  & 0.334  & 0.398 & 0.242   \\
musk             & 1.000 & 0.961 & 0.982 & 1.000          & 1.000 & 0.404   & 1.000 & 0.908 & 1.000 & 0.663 & 1.000 & 1.000 & 0.706 & 0.999    & 0.157 & 1.000 & 0.922 & 0.327      & 1.000 & 1.000  & 0.889  & 1.000 & 1.000   \\
optdigits        & 0.140 & 0.056 & 0.056 & 0.412          & 0.424 & 0.154   & 0.291 & 0.039 & 0.436 & 0.071 & 0.069 & 0.060 & 0.050 & 0.045    & 0.192 & 0.078 & 0.509 & 0.039      & 0.256 & 0.318  & 0.221  & 0.153 & 0.119   \\
pageblocks       & 0.706 & 0.415 & 0.585 & 0.702          & 0.225 & 0.434   & 0.676 & 0.486 & 0.711 & 0.632 & 0.643 & 0.594 & 0.603 & 0.521    & 0.735 & 0.635 & 0.681 & 0.583      & 0.621 & 0.675  & 0.575  & 0.664 & 0.671   \\
pendigits        & 0.512 & 0.309 & 0.415 & 0.857          & 0.423 & 0.588   & 0.970 & 0.372 & 0.786 & 0.132 & 0.518 & 0.386 & 0.117 & 0.093    & 0.146 & 0.334 & 0.664 & 0.145      & 0.611 & 0.919  & 0.592  & 0.484 & 0.283   \\
pima             & 0.721 & 0.691 & 0.648 & 0.695          & 0.759 & 0.737   & 0.754 & 0.594 & 0.684 & 0.686 & 0.720 & 0.712 & 0.565 & 0.598    & 0.534 & 0.652 & 0.786 & 0.712      & 0.712 & 0.797  & 0.696  & 0.680 & 0.563   \\
satellite        & 0.773 & 0.733 & 0.696 & 0.858          & 0.865 & 0.824   & 0.860 & 0.798 & 0.859 & 0.799 & 0.809 & 0.778 & 0.760 & 0.811    & 0.775 & 0.790 & 0.876 & 0.779      & 0.851 & 0.858  & 0.817  & 0.848 & 0.755   \\
satimage-2       & 0.968 & 0.853 & 0.797 & 0.907          & 0.877 & 0.945   & 0.967 & 0.937 & 0.885 & 0.983 & 0.969 & 0.919 & 0.475 & 0.763    & 0.793 & 0.959 & 0.947 & 0.625      & 0.881 & 0.962  & 0.833  & 0.682 & 0.969   \\
shuttle          & 0.968 & 0.981 & 0.952 & 0.464          & 0.975 & 0.986   & 0.979 & 0.557 & 0.998 & 0.909 & 0.977 & 0.963 & 0.660 & 0.980    & 0.134 & 0.602 & 0.997 & 0.517      & 0.979 & 0.981  & 0.994  & 0.940 & 0.244   \\
skin             & 0.695 & 0.297 & 0.305 & 0.492          & 0.534 & 0.646   & 0.982 & 0.530 & 0.617 & 0.624 & 0.663 & 0.364 & 0.504 & 0.430    & 0.656 & 0.422 & 0.325 & 0.747      & 0.764 & 0.948  & 0.969  & 0.691 & 0.571   \\
smtp             & 0.497 & 0.010 & 0.680 & 0.004          & 0.012 & 0.011   & 0.505 & 0.082 & 0.481 & 0.012 & 0.645 & 0.495 & 0.209 & 0.307    & 0.087 & 0.324 & 0.038 & 0.008      & 0.408 & 0.502  & 0.336  & 0.504 & 0.353   \\
spambase         & 0.820 & 0.736 & 0.713 & 0.684          & 0.784 & 0.883   & 0.833 & 0.802 & 0.727 & 0.818 & 0.822 & 0.818 & 0.742 & 0.753    & 0.791 & 0.821 & 0.868 & 0.854      & 0.729 & 0.837  & 0.810  & 0.838 & 0.543   \\
speech           & 0.027 & 0.028 & 0.029 & 0.030          & 0.032 & 0.033   & 0.028 & 0.030 & 0.032 & 0.028 & 0.028 & 0.028 & 0.040 & 0.034    & 0.036 & 0.028 & 0.034 & 0.033      & 0.030 & 0.032  & 0.029  & 0.029 & 0.017   \\
stamps           & 0.622 & 0.564 & 0.490 & 0.656          & 0.523 & 0.588   & 0.717 & 0.572 & 0.648 & 0.417 & 0.649 & 0.588 & 0.465 & 0.426    & 0.285 & 0.496 & 0.795 & 0.524      & 0.647 & 0.825  & 0.728  & 0.577 & 0.613   \\
thyroid          & 0.815 & 0.302 & 0.640 & 0.365          & 0.770 & 0.797   & 0.809 & 0.643 & 0.606 & 0.801 & 0.789 & 0.813 & 0.631 & 0.691    & 0.744 & 0.801 & 0.515 & 0.758      & 0.822 & 0.810  & 0.457  & 0.817 & 0.398   \\
vertebral        & 0.252 & 0.155 & 0.199 & 0.329          & 0.189 & 0.208   & 0.261 & 0.167 & 0.339 & 0.210 & 0.222 & 0.193 & 0.251 & 0.234    & 0.234 & 0.214 & 0.588 & 0.230      & 0.358 & 0.252  & 0.515  & 0.351 & 0.090   \\
vowels           & 0.239 & 0.071 & 0.177 & 0.327          & 0.079 & 0.120   & 0.302 & 0.104 & 0.331 & 0.044 & 0.274 & 0.105 & 0.073 & 0.169    & 0.132 & 0.209 & 0.274 & 0.097      & 0.427 & 0.316  & 0.336  & 0.381 & 0.499   \\
waveform         & 0.225 & 0.099 & 0.074 & 0.287          & 0.090 & 0.105   & 0.270 & 0.078 & 0.307 & 0.078 & 0.109 & 0.084 & 0.061 & 0.115    & 0.201 & 0.089 & 0.186 & 0.251      & 0.093 & 0.279  & 0.196  & 0.100 & 0.059   \\
wbc              & 0.868 & 0.932 & 0.931 & 0.127          & 0.877 & 0.942   & 0.920 & 0.757 & 0.249 & 0.902 & 0.972 & 0.943 & 0.568 & 0.565    & 0.240 & 0.920 & 0.951 & 0.710      & 0.938 & 0.961  & 0.724  & 0.300 & 0.841   \\
wdbc             & 0.757 & 0.838 & 0.610 & 0.937          & 0.778 & 0.720   & 0.820 & 0.548 & 0.936 & 0.553 & 0.874 & 0.821 & 0.309 & 0.843    & 0.122 & 0.788 & 0.956 & 0.775      & 0.843 & 0.905  & 0.921  & 0.688 & 0.804   \\
wilt             & 0.081 & 0.069 & 0.077 & 0.192          & 0.079 & 0.088   & 0.123 & 0.080 & 0.157 & 0.215 & 0.071 & 0.064 & 0.084 & 0.071    & 0.096 & 0.109 & 0.289 & 0.171      & 0.172 & 0.122  & 0.521  & 0.254 & 0.039   \\
wine             & 0.868 & 0.523 & 0.326 & 0.887          & 0.777 & 0.671   & 0.951 & 0.579 & 0.900 & 0.831 & 0.887 & 0.692 & 0.509 & 0.786    & 0.185 & 0.701 & 0.983 & 0.789      & 0.977 & 0.968  & 0.997  & 0.999 & 0.153   \\
wpbc             & 0.448 & 0.382 & 0.358 & 0.410          & 0.426 & 0.407   & 0.461 & 0.383 & 0.412 & 0.452 & 0.409 & 0.400 & 0.372 & 0.749    & 0.360 & 0.389 & 0.893 & 0.455      & 0.546 & 0.690  & 0.658  & 0.604 & 0.244   \\
yeast            & 0.507 & 0.468 & 0.494 & 0.499          & 0.498 & 0.468   & 0.483 & 0.490 & 0.489 & 0.457 & 0.480 & 0.468 & 0.518 & 0.492    & 0.498 & 0.508 & 0.496 & 0.470      & 0.511 & 0.481  & 0.511  & 0.497 & 0.323   \\
CIFAR10          & 0.197 & 0.121 & 0.126 & 0.222          & 0.140 & 0.165   & 0.196 & 0.169 & 0.222 & 0.159 & 0.194 & 0.192 & 0.120 & 0.140    & 0.124 & 0.194 & 0.174 & 0.159      & 0.196 & 0.199  & 0.167  & 0.197 & 0.693   \\
MNIST-C          & 0.425 & 0.095 & 0.095 & 0.522          & 0.216 & 0.328   & 0.462 & 0.326 & 0.519 & 0.258 & 0.416 & 0.403 & 0.234 & 0.314    & 0.269 & 0.412 & 0.515 & 0.341      & 0.418 & 0.474  & 0.441  & 0.472 & 0.506   \\
MVTec-AD         & 0.749 & 0.378 & 0.378 & 0.758          & 0.676 & 0.700   & 0.758 & 0.657 & 0.758 & 0.805 & 0.730 & 0.721 & 0.581 & 0.838    & 0.593 & 0.726 & 0.895 & 0.679      & 0.737 & 0.829  & 0.829  & 0.851 & 0.806   \\
SVHN             & 0.151 & 0.095 & 0.095 & 0.161          & 0.120 & 0.139   & 0.153 & 0.127 & 0.160 & 0.128 & 0.150 & 0.149 & 0.114 & 0.124    & 0.112 & 0.149 & 0.156 & 0.142      & 0.151 & 0.155  & 0.142  & 0.155 & 0.077   \\
mnist            & 0.665 & 0.169 & 0.169 & 0.693          & 0.222 & 0.542   & 0.727 & 0.341 & 0.710 & 0.558 & 0.662 & 0.650 & 0.461 & 0.460    & 0.597 & 0.651 & 0.685 & 0.552      & 0.624 & 0.737  & 0.561  & 0.563 & 0.991   \\
FashionMNIST     & 0.578 & 0.095 & 0.095 & 0.639          & 0.349 & 0.447   & 0.592 & 0.469 & 0.636 & 0.374 & 0.565 & 0.562 & 0.297 & 0.451    & 0.296 & 0.566 & 0.631 & 0.468      & 0.571 & 0.598  & 0.537  & 0.550 & 0.989   \\
20news           & 0.126 & 0.111 & 0.113 & 0.150          & 0.111 & 0.116   & 0.135 & 0.112 & 0.150 & 0.155 & 0.118 & 0.113 & 0.102 & 0.129    & 0.120 & 0.115 & 0.146 & 0.106      & 0.115 & 0.156  & 0.141  & 0.173 & 0.241   \\
agnews           & 0.138 & 0.111 & 0.109 & 0.259          & 0.112 & 0.119   & 0.167 & 0.121 & 0.259 & 0.146 & 0.128 & 0.116 & 0.102 & 0.102    & 0.097 & 0.124 & 0.154 & 0.097      & 0.119 & 0.174  & 0.128  & 0.192 & 0.367   \\
amazon           & 0.115 & 0.112 & 0.104 & 0.111          & 0.111 & 0.111   & 0.117 & 0.102 & 0.110 & 0.117 & 0.111 & 0.107 & 0.095 & 0.102    & 0.095 & 0.109 & 0.102 & 0.096      & 0.108 & 0.117  & 0.102  & 0.112 & 0.061   \\
imdb             & 0.090 & 0.093 & 0.085 & 0.090          & 0.090 & 0.090   & 0.089 & 0.087 & 0.090 & 0.095 & 0.089 & 0.087 & 0.092 & 0.097    & 0.099 & 0.088 & 0.102 & 0.095      & 0.087 & 0.090  & 0.101  & 0.089 & 0.060   \\
yelp             & 0.137 & 0.133 & 0.119 & 0.161          & 0.130 & 0.132   & 0.160 & 0.118 & 0.161 & 0.138 & 0.134 & 0.128 & 0.093 & 0.100    & 0.101 & 0.131 & 0.104 & 0.107      & 0.128 & 0.164  & 0.123  & 0.130 & 0.057   \\
\midrule
\textbf{Average} & 0.499 & 0.388 & 0.401 & 0.469          & 0.419 & 0.438   & 0.562 & 0.383 & 0.521 & 0.439 & 0.504 & 0.470 & 0.356 & 0.444    & 0.334 & 0.440 & 0.557 & 0.434      & 0.524 & 0.571  & 0.545  & 0.520 & 0.440  \\ 
\bottomrule
\end{tabular}
\caption{PR AUC for the semi-supervised setting on ADBench}
\label{tab:adbench4}
\end{sidewaystable}

\clearpage
\subsection{Ablation studies}
\vspace{0.2cm}
\textbf{The optimal loss usage percentage} \hspace{0.2cm} We examine the optimal amount of loss truncation required when implementing ALTBI. The table shows the averaged AUC on \texttt{ADBench} datasets for various loss usage percentages. We can identify that using 92 percent of loss values performs best in our method.
\begin{table}[h!]
\renewcommand\thetable{B.6}
\centering
\setlength{\tabcolsep}{2mm} 
\fontsize{9pt}{11.6pt}\selectfont 
\begin{tabular}{l|cccccc}
\hline
 $\boldsymbol{\rho}$ \hspace{0.8cm} & \textbf{0.90}        & \textbf{0.92}  & \textbf{0.94}      & \textbf{0.96}       & \textbf{0.98}       & \textbf{1.00}          \\

\hline
\textbf{AUC}             & 0.759 & \textbf{0.762} & 0.759 & 0.758 & 0.756 & 0.755 \\
\hline
\end{tabular}
\caption{Averaged results of training AUC scores with various values of $\rho$.}
\label{tab:ablation_percent}
\end{table}\\
\textbf{The increase in mini-batch sizes} \hspace{0.2cm} We investigate how much to increase the mini-batch size at each update for optimal performance. There was no significant difference in scores with varying $\gamma$ values, but the highest score was achieved with a $\gamma$ of 1.03. Therefore, we chose this value for our experiments.
\begin{table}[h!]
\renewcommand\thetable{B.7}
\centering
\setlength{\tabcolsep}{1.7mm} 
\fontsize{10pt}{11.6pt}\selectfont 
\begin{tabular}{l|cccccccc}
\hline
$\boldsymbol{\gamma}$ \hspace{0.8cm} & \textbf{1.00}      & \textbf{1.01}   & \textbf{1.02}   & \textbf{1.03}   & \textbf{1.04}   & \textbf{1.05}   & \textbf{1.07}   & \textbf{1.1}    \\
\hline
\textbf{AUC}      & 0.761 & 0.760 & 0.761 & \textbf{0.762} & 0.761 & 0.762 & 0.761 & 0.761\\
\hline
\end{tabular}
\caption{Averaged results of training AUC scores with various values of $\gamma$}
\label{tab:ablation_gamma}
\end{table}\\
\textbf{Number of samples used in the IWAE} \hspace{0.2cm}
Table \ref{tab:ablation_k} shows the averaged AUC values on 57 \texttt{ADBench} datasets with various numbers of samples in the IWAE ranging from 1 to 100. It should be noted that K=1 is equivalent to the original VAE. We observe that as the value of K increases, the scores decline and then stabilize. Therefore, we choose K=2 in our experiments.
\begin{table}[h!]
\renewcommand\thetable{B.8}
\centering
\setlength{\tabcolsep}{1.7mm} 
\fontsize{10pt}{11.6pt}\selectfont 
\begin{tabular}{l|cccccccc}
\hline
\textbf{K} \hspace{0.8cm}   & \textbf{1}     & \textbf{2}     & \textbf{5}     & \textbf{10}    & \textbf{20}    & \textbf{50}    & \textbf{70}    & \textbf{100}   \\
\hline
\textbf{AUC} & 0.760 & \textbf{0.762} & 0.755 & 0.760 & 0.759 & 0.758 & 0.753 & 0.753\\
\hline
\end{tabular}
\caption{Averaged results of training AUC scores with various values of K}
\label{tab:ablation_k}
\end{table}\\
\textbf{Learning Schedule} \hspace{0.2cm}
We examine the performance of ALTBI with respect to the learning rate. We use the Adam optimizer with various learning rates ranging from 1e-4 to 1e-1, and the table \ref{tab:ablation_lr} compares the averaged AUC on \texttt{ADBench} datasets. We observe that when the learning rate is larger than 1e-03, the performance of the model deteriorates and stabilizes. For this reason, we set the learning rate to 1e-03 in our experiments.
\begin{table}[h!]
\renewcommand\thetable{B.9}
\centering
\setlength{\tabcolsep}{1.7mm} 
\fontsize{10pt}{11.6pt}\selectfont 
\begin{tabular}{l|cccccccccc}
\hline
\textbf{Learning rate} \hspace{0.2cm}   & \textbf{1e-04} & \textbf{2.5e-04} & \textbf{5e-04} & \textbf{1e-03} & \textbf{2.5e-03} & \textbf{5e-03} & \textbf{1e-02} & \textbf{2.5e-02} & \textbf{5e-02} & \textbf{1e-01} \\
\hline
\textbf{AUC} & 0.712    & 0.743    & 0.756    & \textbf{0.762}    & 0.744    & 0.738    & 0.735    & 0.737    & 0.739    & 0.738   \\
\hline
\end{tabular}
\caption{Averaged results of training AUC scores with various values of learning rate}
\label{tab:ablation_lr}
\end{table}\\

\newpage
\noindent
\textbf{Iteration for ensembling within a single model} \hspace{0.2cm}
We evaluate the best $T_1$ and $T_2$ for ensembling within a single model. We compare  three values of $T_1$ and four values of $T_2 - T_1$, and the table \ref{tab:ablation_ens} shows the averaged results on \texttt{ADBench} datasets. We can see that low values of $T_1$ and $T_2$ result in lower scores, so we set $T_1$ to 70 and $T_2$ to 90.
\begin{table}[h!]
\renewcommand\thetable{B.10}
\centering
\setlength{\tabcolsep}{1.7mm} 
\fontsize{9pt}{11.6pt}\selectfont 
\begin{tabular}{l|ccc}
\hline
            & \multicolumn{3}{c}{$\textbf{T}_\textbf{1}$} \\
            \hline
$\textbf{T}_\textbf{2}-\textbf{T}_\textbf{2}$ & \textbf{50}     & \textbf{60}     & \textbf{70}     \\
\hline
\textbf{5}           & 0.757  & 0.760  & 0.760  \\
\textbf{10}          & 0.758  & 0.761  & 0.760  \\
\textbf{20}          & 0.759  & 0.760  & \textbf{0.762}  \\
\textbf{30}          & 0.760  & 0.762  & 0.760 \\
\hline
\end{tabular}
\caption{Averaged results of training AUC scores with various values of iteration for ensembling}
\label{tab:ablation_ens}
\end{table}

\noindent
\textbf{ALTBI for addressing SSOD} \hspace{0.2cm} 
We compare ALTBI with other methods in SSOD tasks. 
First, we split inliers into two sets with ratios of $7:3$ and use the first partition as a training dataset, and regard the rest inliers and outliers as the test dataset.  

\begin{figure}[h!]
\renewcommand\thefigure{B.1}
\centering
\includegraphics[width=0.5\columnwidth]{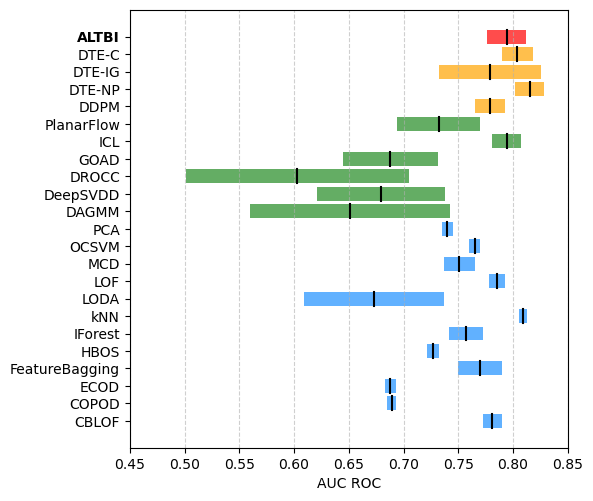} 
\caption{Test AUC ROC means and standard deviation on the 57 datasets from ADBench over five different seeds in semi-supervised setting. Color scheme: red (IM-based), orange (diffusion-based), green (deep-learning-based), blue (machine-learning-based).}
\label{fig:performace_results_ssod}
\end{figure}

Table \ref{tab:adbench3}\&\ref{tab:adbench4}, and Figure \ref{fig:performace_results_ssod} show the AUC and PRAUC results of the test datasets. 
We can observe that ALTBI is still competitive in addressing SSOD tasks.  

\clearpage
\subsection{Further discussions: Robustness of ALTBI in DP}
DP-SGD \citep{abadi2016deep} is a variant of SGD used for updating parameters while ensuring differential privacy (DP, \citet{10.1007/11787006_1}) for the model. 
For each sample \(\boldsymbol{x}\), with its corresponding per-sample loss \(\tilde{l}(\theta;\boldsymbol{x})\), we compute the gradient vector \(\nabla_{\theta}l(\theta;\boldsymbol{x})\). 
This gradient is then clipped using a specified positive constant \(C > 0\) and combined with Gaussian noise drawn from \(\mathcal{N}(0,\sigma^2C^2 I)\) to produce a modified gradient:
\begin{align*}
{\nabla}_{\theta}^{\text{DP}}\tilde{l}(\theta;\boldsymbol{x}) = \frac{{\nabla}_{\theta}\tilde{l}(\theta;\boldsymbol{x})}{\max\left(1,\frac{\|{\nabla}_{\theta}\tilde{l}(\theta;\boldsymbol{x})\|_2}{C}\right)} + \mathcal{N}(0,\sigma^2C^2 I),
\end{align*}
where \(\sigma > 0\) is another pre-specified constant. 
In practice, we set $(C,\sigma)=(10,0.7)$ to carry out DP experiments. 
The parameters \(\theta\) is then updated using this modified gradient \(\tilde{l}\) with the conventional SGD method or its variants such as Adam \citep{kingma2014adam}.

For a given loss function to be compatible with DP-SGD, the loss function must be separable across mini-batch samples. 
However, ALTBI uses a threshold that is calculated based on the current per-sample loss values. 
Due to this, DP-SGD cannot be directly applied to ALTBI, as the truncated loss function is not separable.
This issue can be easily addressed by calculating the threshold using the per-sample loss values from the previous update, i.e., using $\tau_{t-1}$ instead of $\tau_{t}$.

We consider two versions of ALTBI: one that applies mini-batch increment and truncated loss, i.e., $(\gamma,\rho)=(1.03,0.92)$, and one that does not,i.e., $(\gamma,\rho)=(1.0,1.0)$. 
As a measure of DP, we adopt $(\epsilon,\delta)$-DP, which is the standard measure to assess differential privacy. 
With a fixed \(\delta = 1 \times 10^{-5}\), we train the models using a DP-SGD algorithm until the privacy budget \(\epsilon\) does not exceed 10. 
To implement DP-SGD, we use the \texttt{Opacus} library \citep{opacus} in Python. 

Similar to the SSOD analysis, we split the entire dataset into two parts with a 7:3 ratio, treating the larger portion as training data and the remaining portion as test data. 
The test AUC results for 20 tabular datasets are presented in Table \ref{tab:dp_results}.
We observe that using mini-batch increment and the truncated loss function results in less degradation in averaged AUC performance (from 0.759 to 0.750) compared to not using these techniques (from 0.729 to 0.651). 
This indicates that our method provides robustness in performance when applying DP.


\begin{table}[h!]
\renewcommand\thetable{B.11}
\centering
\setlength{\tabcolsep}{1.7mm} 
\fontsize{10pt}{11.6pt}\selectfont 
\begin{tabular}{l|cc|cc}
\hline
                 & \multicolumn{2}{l|}{\textbf{($\boldsymbol{\gamma},\boldsymbol{\rho}$) = (1.03,0.92)}} & \multicolumn{2}{l}{\textbf{($\boldsymbol{\gamma},\boldsymbol{\rho}$) = (1.0,1.0)}} \\ \hline
\textbf{Data}    & \textbf{w/o DP}      & \textbf{w/ DP}     & \textbf{w/o DP}    & \textbf{w/ DP}    \\ \hline
breastw          & 0.725                & 0.981              & 0.981              & 0.969             \\
cardio           & 0.542                & 0.927              & 0.802              & 0.820             \\
cardiotocography & 0.803                & 0.669              & 0.566              & 0.623             \\
celeba           & 0.662                & 0.809              & 0.739              & 0.711             \\
census           & 0.901                & 0.569              & 0.665              & 0.464             \\
fault            & 0.923                & 0.588              & 0.684              & 0.557             \\
fraud            & 0.771                & 0.928              & 0.943              & 0.658             \\
landsat          & 0.744                & 0.368              & 0.519              & 0.296             \\
magic.gamma      & 0.806                & 0.780              & 0.794              & 0.722             \\
musk             & 0.684                & 0.439              & 1.000              & 0.393             \\
optdigits        & 0.868                & 0.552              & 0.629              & 0.520             \\
pageblocks       & 0.954                & 0.790              & 0.882              & 0.776             \\
pima             & 0.760                & 0.716              & 0.715              & 0.683             \\
satimage-2       & 0.982                & 0.952              & 0.998              & 0.865             \\
skin             & 0.846                & 0.889              & 0.804              & 0.859             \\
spambase         & 0.477                & 0.722              & 0.545              & 0.437             \\
speech           & 0.883                & 0.581              & 0.474              & 0.487             \\
stamps           & 0.947                & 0.849              & 0.876              & 0.805             \\
wine             & 0.495                & 0.914              & 0.886              & 0.907             \\
wpbc             & 0.411                & 0.559              & 0.503              & 0.478             \\ \hline
\textbf{Average} & 0.759                & 0.729              & 0.750              & 0.651    \\
\hline
\end{tabular}
\caption{
The average test AUC results for 20 datasets with and without DP applied under the conditions of $\gamma=1.03, \rho=0.92$ and $\gamma=1.0, \rho=1.0$}
\label{tab:dp_results}
\end{table}


\end{document}